\documentclass[letterpaper,journal]{IEEEtran}

\usepackage{amsmath,amssymb}

\usepackage{graphicx}
\usepackage{booktabs}
\usepackage{etoolbox}
\usepackage[caption=false,font=footnotesize]{subfig}

\usepackage{cite}

\usepackage[ruled,linesnumbered]{algorithm2e}
\SetAlgoLined
\SetAlgoShortEnd

\usepackage[
    colorlinks=true,
    linkcolor=blue,
    citecolor=blue,
    urlcolor=blue
]{hyperref}

\begin{document}





\title{Video2Track: Steerable Adversarial Scenario
Generation from Real-World Interaction Videos
to Proving Grounds for Automated
Driving System Testing}


\author{
Mengjie Tian, Xinrui Zhang, Tianyu Li, Peizhi Zhang,
Guirong Zhuo, Haojie Feng, Ziyi Yin,\\
Qixiang~Zhang, and Lu Xiong%
\thanks{All authors except Qixiang Zhang are with
Tongji University, Shanghai, China.
Qixiang Zhang is with Southeast University,
Nanjing, China.}%
\thanks{Mengjie Tian and Xinrui Zhang contributed
equally to this work.
Corresponding author: Peizhi Zhang
(zhangpeizhitom@126.com).}%
}

\maketitle


\begin{abstract}
Physical proving-ground testing enables
controlled and repeatable validation of
automated driving systems (ADSs).
However, existing approaches often rely
on manually designed reference routes
and scripted vehicle behaviors, limiting
deployment efficiency, real-world
interaction reproduction, and
interaction risk steerability.
To address these limitations, we propose
Video2Track, a framework that converts
real-world driving videos into steerable
adversarial scenarios for proving-ground
testing.
Video2Track extracts interaction semantics
using a vision-language model, retrieves
compatible maneuver pairs on the target
road network, and generates reference
trajectories through interaction-guided
diffusion.
Game-based online refinement adjusts
background vehicle (BV) trajectories based
on real-time feedback from the vehicle
under test (VUT) to regulate interaction
risk toward a prescribed anticipated
post-encroachment time (APET).
Video2Track achieved an interaction
consistency rate of 82.1\% and a scenario
generation success rate of 78.6\%.
In proving-ground tests, the mean minimum
APET was within 0.5~s of each of the four
target values.
Video2Track also achieved lower mean
minimum APET values than the evaluated
baselines in both cut-in and unprotected
left-turn scenarios.
These results demonstrate the feasibility
of transferring real-world driving interactions
to executable proving-ground scenarios,
enabling faithful interaction reproduction
and steerable adversarial testing on an
Internet of Things (IoT)-enabled
5G cloud-controlled platform.
\end{abstract}


\section{Introduction}\label{sec:1}

\IEEEPARstart{A}{utomated} driving systems
(ADSs) operate in complex, dynamic, and
highly interactive traffic environments,
making comprehensive safety validation
essential before real-world
deployment~\cite{Zhangzq,11457031,zhang2025real}.
Such validation provides evidence for
safety cases supporting ADS safety
assurance throughout their
lifecycle~\cite{WANG202417}.
Public-road testing exposes ADSs to
realistic traffic interactions but
faces safety risks, limited exposure
to safety-critical
events~\cite{zhangpeixing2023,liu2024curse},
and poor scenario reproducibility.
As an important component of physical
safety validation, proving-ground testing
offers a safe, controllable, and repeatable
environment for evaluating ADS behavior
under real vehicle dynamics and actuator
constraints.

However, conventional proving-ground
testing often relies on standardized
protocols, manual configuration of
testing locations and vehicle routes,
and predefined vehicle
trajectories~\cite{yang2021cavtest}.
These approaches can result in scripted
background vehicle (BV) behaviors that
insufficiently capture dynamic interactions
with the vehicle under test
(VUT)~\cite{ma2024evolving,li2026mrisce},
limiting interaction fidelity,
steerability, and testing efficiency.

Recent advances in virtual--real
integration~\cite{zhu2026Iotj,feng2020safety}
and cloud-controlled testing
systems~\cite{zhang2025real,tian2025cloud}
have enhanced the capabilities of
physical proving grounds.
In parallel, mechanism-based and
data-driven methods have been developed
to generate realistic and challenging
traffic scenarios~\cite{zhang2025testing,
liu2025isfm4sim,wu2025evolving,
wang2022autonomous}.
The feasibility of transferring
naturalistic driving scenarios to
proving grounds has also been explored
through trajectory and road-geometry
matching~\cite{chen2021how}.
More recently, vision-language models
have facilitated semantic understanding
of driving scenarios, while diffusion
models have advanced controllable
trajectory generation~\cite{xu2025diffscene,
lu2025omnitester,zhang2025drivegen}.
Chat2Scenic~\cite{chat2scenic} further
employs retrieval-augmented generation
to translate natural-language scenario
descriptions into executable Scenic
scripts for simulation.
Video-driven approaches have also explored
extracting interaction semantics from
real-world driving videos to construct
corresponding test
scenarios~\cite{lu2025omnitester,liuvlm}.

Nevertheless, transferring video-derived
interactions to physical proving grounds
remains challenging.
Even with compatible road topologies,
differences in road geometry and scale
complicate the automatic identification
of suitable testing locations and
maneuver pairs.
The resulting scenarios must reproduce
the source interaction behaviors and
risk characteristics while satisfying
vehicle motion constraints.
Moreover, real-time VUT responses may
alter interaction timing and risk,
requiring online refinement.
Addressing these challenges requires
three key capabilities for video-driven
proving-ground testing: (i) \emph{fidelity}, reproducing
source vehicle maneuvers, interaction
behaviors, and risk characteristics
through compatible spatial configurations
on the fixed proving-ground topology;
(ii) \emph{steerability}, actively
adapting BV behavior to real-time
responses of the black-box VUT
to steer interaction risk toward
specified targets;
and (iii) \emph{adversariality},
generating challenging interactions
for ADS stress testing.

In this paper, we propose Video2Track,
an integrated framework for transferring
real-world driving interactions from
videos into executable adversarial
scenarios on physical proving grounds.
The framework consists of two main
modules: scenario generation and
online interaction refinement.
In the first module, video-derived
interaction semantics are grounded
onto the target road network through
maneuver-pair retrieval, followed by
interaction-guided diffusion to generate
initial reference trajectories consistent
with the intended maneuvers and
passing orders.
In the second module, game-based online
interaction refinement adjusts BV
trajectories in response to real-time
VUT behavior, enabling steerable
interaction risk during physical testing.
A 5G cloud-controlled platform supports
vehicle-state feedback and closed-loop
BV trajectory execution.

The main contributions are summarized
as follows.

\begin{enumerate}

    \item Video2Track is proposed as an integrated
    framework for transferring real-world driving
    interactions from videos to executable
    proving-ground scenarios on a 5G
    cloud-controlled platform, enabling
    faithful interaction reproduction and steerable
    adversarial testing.

    \item An interaction-aware scenario generation
    method is proposed, combining maneuver-pair
    retrieval and interaction-guided diffusion
    to generate reference trajectories that
    preserve vehicle maneuvers and passing
    orders on the target road network.

    \item A game-based online interaction refinement
    method is developed to adjust BV trajectories
    using generated references and real-time
    VUT feedback, enabling interaction risk
    regulation toward prescribed APET targets
    during proving-ground testing.
\end{enumerate}

The remainder of this paper is organized
as follows.
Section~\ref{sec:2} reviews related work.
Section~\ref{sec:3} presents the
Video2Track framework and its key methods.
Section~\ref{sec:4} reports the
proving-ground experimental results
and analysis.
Section~\ref{sec:5} concludes the paper
and discusses future work.

\section{Related Work}\label{sec:2}

\subsection{Scenario Generation Methods}

Scenario generation methods can be broadly
categorized into rule- and knowledge-based
approaches and data-driven approaches.
Rule- and knowledge-based methods generate
scenarios through predefined templates,
parameter sampling, and explicit
constraints~\cite{xia2018test,
fremont2019scenic,winner2018pegasus},
offering interpretability and
reproducibility but often requiring
manual scenario specification.
In contrast, data-driven methods learn
traffic patterns from naturalistic data
to improve scenario realism and
diversity~\cite{dingzhaosurvy}.
For instance, TrafficGen~\cite{trafficgen}
generates vehicle configurations and
trajectories autoregressively,
while CTG~\cite{ctg2023} employs
logic-guided diffusion for controllable
traffic generation.
DiffScene~\cite{xu2025diffscene}
further incorporates adversarial objectives
to generate realistic safety-critical
scenarios.
Despite these advances, reproducing
the interaction semantics of specific
real-world events remains challenging,
particularly when scenarios must be
deployed on fixed physical
proving-ground layouts.

Recent studies have incorporated
vision-language models (VLMs),
large language models (LLMs), and
generative models to improve semantic
controllability.
ScriptGPT~\cite{miao2024dashcam}
translates dashcam videos into Scenic
scripts for simulation in CARLA.
OmniTester~\cite{lu2025omnitester}
generates testing scenarios from
user-specified requirements, while
DriveGen~\cite{zhang2025drivegen}
uses VLM-generated waypoint goals
to guide diffusion-based traffic generation.
ChatScene~\cite{chatscene} retrieves
Scenic code snippets for road geometry,
vehicle placement, and adversarial
behaviors to instantiate scenarios
on CARLA maps.
Chat2Scenic~\cite{chat2scenic}
further employs retrieval-augmented
generation to construct Scenic scripts
with map configurations, spatial
relations, and vehicle behaviors.
Another video-driven
approach~\cite{liuvlm} generates
executable Scenic code from driving
videos for simulation-based testing.
Although these methods advance
automated scenario generation,
transferring video-derived interactions
to physical proving grounds additionally
requires compatible spatial configurations,
executable trajectories, and preservation
of the intended vehicle interactions.

\subsection{Scenario-Based Adversarial Testing for ADS}

Standardized scenario-based testing
is widely used in ADS safety
evaluation~\cite{kullgren2010comparison}.
Euro NCAP specifies standardized
testing scenarios and procedures
for vehicle safety assessment~\cite{euroncap2024aeb},
while ISO 34502:2022~\cite{iso2023road}
provides a framework for scenario-based
ADS safety evaluation.
Conventional proving-ground tests
often employ predefined trajectory
replay or rule-based behavior triggers
to control BVs.
Although these approaches support
reproducibility and comparability,
their scripted behaviors cannot
continuously adapt to dynamic VUT
responses, limiting the control of
interaction timing and risk.

Search-based and reinforcement learning
(RL)-based methods have been investigated
for adversarial scenario testing.
Search-based approaches explore scenario
parameter spaces to identify
safety-critical conditions~\cite{sun2021scenario},
while RL-based methods learn adversarial
BV behaviors to expose safety-critical
events~\cite{liTITS2026,feng2023dense}.
For example, D2RL~\cite{feng2023dense}
leverages naturalistic driving data
to accelerate safety evaluation and
has been validated on physical test
tracks using augmented-reality traffic
environments.
However, these studies primarily focus
on safety-critical scenario discovery
and adversarial policy learning,
rather than transferring specific
video-derived interactions to
physical proving grounds.

Game-theoretic methods model
interdependent vehicle behaviors.
The method in~\cite{ma2024evolving}
combines level-$k$ reasoning with
reinforcement learning to generate
evolving interaction scenarios.
TroubleMaker~\cite{zhang2025real}
employs Stackelberg optimization
to adapt BV behavior to VUT responses
during physical testing.
However, its test configurations and reference
routes are predefined.
Automated transfer of real-world driving
interactions into executable proving-ground
scenarios, integrated with online adversarial
refinement, remains insufficiently explored.


\section{Methodology}\label{sec:3}

This section presents the technical details
of Video2Track, a framework for transferring
real-world driving interactions from videos
into steerable adversarial scenarios
for physical proving-ground testing of ADSs.

\subsection{Framework Overview}
\label{sec:3.1}

\begin{figure*}[!t]
    \centering
    \includegraphics[width=\textwidth]
    {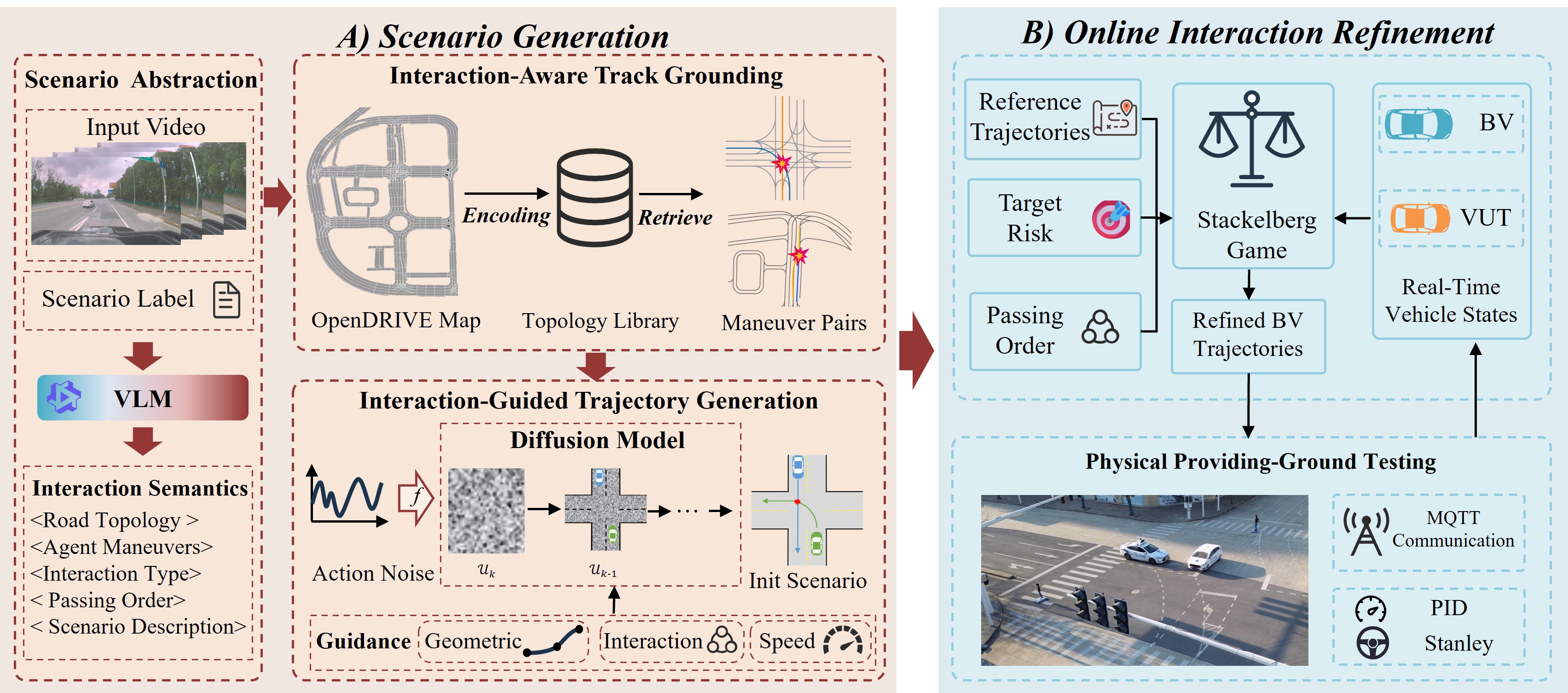}
    \caption{Overview of the proposed
    Video2Track framework.
    (A) Scenario generation extracts
    interaction semantics from real-world
    driving videos, retrieves compatible
    maneuver pairs on the target
    proving-ground road network, and
    generates initial reference trajectories
    through interaction-guided diffusion.
    (B) Online interaction refinement
    employs game-based optimization
    and real-time VUT feedback to refine
    BV trajectories, enabling steerable
    interaction risk during physical
    proving-ground testing.}
    \label{FIG:framework}
\end{figure*}

As illustrated in Fig.~\ref{FIG:framework},
Video2Track consists of two main modules:
scenario generation and online interaction
refinement.
In the first module, a VLM extracts
structured interaction semantics from
real-world driving videos, including
road topology, vehicle maneuvers,
interaction types, and passing orders.
These semantics are grounded onto
the target proving-ground road network
through interaction-aware maneuver-pair
retrieval, which determines reference
routes and spatial anchors.
Interaction-guided diffusion then
generates initial reference trajectories
consistent with the intended maneuvers
and passing orders. In the second module, game-based online
interaction refinement adapts BV trajectories
based on the initial references and
real-time VUT feedback.
The refined trajectories are executed
during physical proving-ground testing,
enabling interaction risk to be steered
toward target APET values.

\subsection{Scenario Generation}
\label{sec:3.2}

\subsubsection{Scenario Abstraction}
\label{sec:3.2.1}

Faithful reproduction of real-world driving
interactions on proving grounds requires
an understanding of the underlying
scenario semantics and interaction relationships.
To this end, we employ GPT-6 to extract
structured interaction semantics from
driving videos in our publicly released
\textit{VideoScenario}
dataset~\cite{videoscenario_hf}.
Following the training-free Scene-CoT
method~\cite{cannons2026from},
the videos are analyzed through
multi-turn prompting to identify
road topology, vehicle maneuvers,
interaction types, and passing orders.
The extracted interaction semantics
are represented as
\begin{equation}
\mathbf{s}
=
\mathcal{F}(V,P)
=
\big(
\Psi,e,\{b_i,\phi_i,\gamma_i\}_{i=1}^{N},
\mathcal{E}
\big)
\label{eq:scenario_semantics}
\end{equation}
where $V$ and $P$ denote the input video
and structured prompts, respectively.
$\Psi$ denotes the road topology semantics,
while $e$ and $b_i$ describe the ego
vehicle and the $i$-th BV, including
their road-relative positions and maneuvers.
$\phi_i$ and $\gamma_i$ denote the
interaction type and passing order,
respectively.
$N$ is the number of BVs,
and $\mathcal{E}$ denotes the
scenario-level textual description.
The critical BV and target APET
are specified by external scenario labels,
rather than inferred by the VLM.

\subsubsection{Interaction-Aware Track Grounding}
\label{sec:3.2.2}

\begin{figure}[t]
    \centering
    \includegraphics[width=\columnwidth]
    {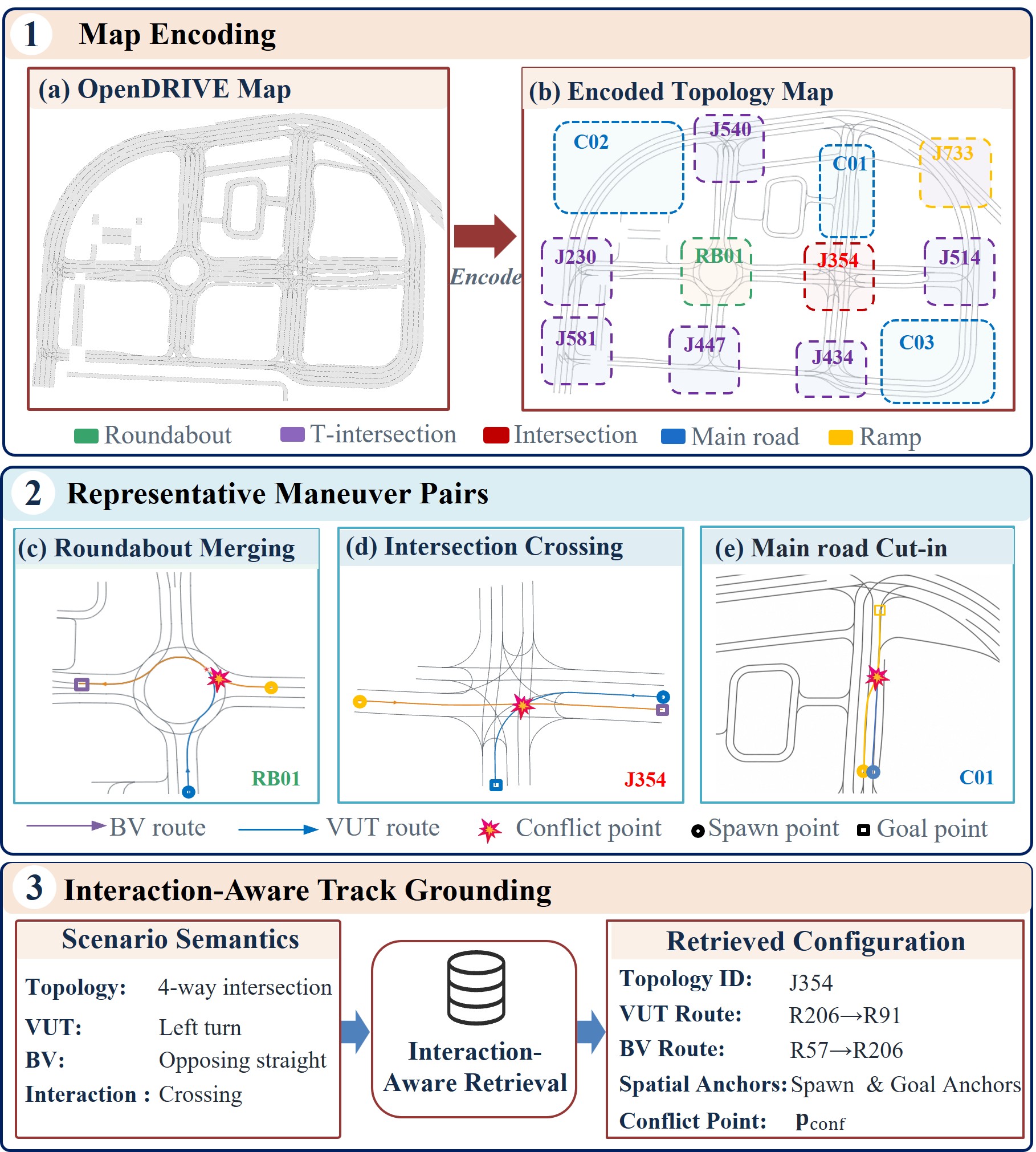}
    \caption{OpenDRIVE-based topology encoding
    and interaction-aware track grounding.
    (a) Proving-ground road network.
    (b) Encoded topology regions.
    (c)--(e) Representative maneuver pairs
    with vehicle routes and conflict points.
    The bottom panel illustrates the retrieval
    of a compatible spatial configuration
    from video-derived interaction semantics.}
    \label{fig:track_grounding}
\end{figure}

Automatically grounding video-derived
road topology semantics and vehicle
interactions onto compatible spatial
configurations on a proving ground
is challenging.
We employ interaction-aware track grounding
based on a topology library $\mathcal{D}$
constructed from the OpenDRIVE map.
The library adopts a three-level
representation: lane segments, maneuvers,
and maneuver pairs.
Lane segments encode road geometry and
directed connectivity, while maneuvers
represent feasible routes connecting
incoming and outgoing lanes.
Maneuver pairs characterize the geometric
and topological relationships between
two routes, including interaction types
and conflict locations.
This representation enables joint
route matching while preserving
the intended interaction relationships.
The topology library and representative
maneuver pairs are illustrated in
Fig.~\ref{fig:track_grounding}.

Given the video-derived road topology
semantics $\Psi$, the ego and BV descriptions
$e$ and $b$, and their interaction type
$\phi$, the target maneuver pair is retrieved as
\begin{equation}
\mathcal{M}^{*}
=
\operatorname{Retrieve}
\left(
\Psi,e,b,\phi;\mathcal{D}
\right)
\label{eq:topology_retrieval}
\end{equation}
where $\mathcal{M}^{*}$ denotes the
retrieved maneuver pair.

Candidate maneuver pairs are first filtered
according to road topology, directed lane
connectivity, vehicle maneuvers, and
interaction type.
The feasible candidates are then ranked
by their consistency with the video-derived
lane layout, relative approach directions,
and entry-exit relationships.
The highest-ranked pair is selected,
with predefined proving-ground priorities
used to resolve ties.

The retrieved maneuver pair determines
a spatial configuration consistent with
the intended vehicle maneuvers and
interaction type.
Reference routes are derived from
lane-centerline geometry, and the
conflict point $\mathbf{p}_{\mathrm{conf}}$
is determined from the interaction
geometry of the paired routes.
Spawn and goal anchors are determined
based on the road-relative entry
positions and intended exit regions.
The resulting spatial configuration
is defined as
\begin{equation}
\mathcal{C}_{\mathrm{map}}
=
\big(
\{\mathcal{R}_i,\mathbf{p}_{\mathrm{spawn}}^i,
\mathbf{p}_{\mathrm{goal}}^i\}
_{i\in\{\mathrm{ego},\mathrm{BV}\}},
\mathbf{p}_{\mathrm{conf}}
\big)
\label{eq:spatial_mapping}
\end{equation}
where $\mathcal{R}_i$ denotes the reference
route, and $\mathbf{p}_{\mathrm{spawn}}^i$
and $\mathbf{p}_{\mathrm{goal}}^i$ denote
the spawn and goal anchors, respectively.
These spatial constraints, together with
the prescribed passing order $\gamma$,
guide subsequent trajectory generation.

\subsubsection{Interaction-Guided Trajectory Generation}
\label{sec:3.2.3}

Given the extracted interaction semantics
and grounded spatial constraints,
we employ an interaction-guided diffusion model
to generate initial reference trajectories
for the ego vehicle and BV.
Geometric and interaction guidance encourage
consistency with the target road topology
and intended interaction relationships.

Let $\mathcal{I}=\{\mathrm{E},\mathrm{B}\}$
denote the ego vehicle and BV.
Their joint state and control trajectories
are denoted by
$\mathcal{X}=\{X^i\}_{i\in\mathcal{I}}$
and $\mathcal{A}=\{A^i\}_{i\in\mathcal{I}}$,
respectively.
For each vehicle $i\in\mathcal{I}$,
the state and control vectors are
\begin{equation}
\begin{aligned}
x_t^i
&=
[p_{x,t}^i,p_{y,t}^i,\theta_t^i,v_t^i]^\top,\\
u_t^i
&=
[a_t^i,\omega_t^i]^\top
\end{aligned}
\end{equation}
where the variables represent planar
position, heading, speed, longitudinal
acceleration, and yaw rate.
Vehicle motion follows the discrete-time
dynamics defined in our prior
work~\cite{tian2025cloud}:
\begin{equation}
x_{t+1}^i=f(x_t^i,u_t^i).
\label{eq:vehicle_dynamics}
\end{equation}
The joint trajectories are reconstructed as
$\mathcal{X}=\operatorname{Rollout}
(\mathcal{X}_{\mathrm{init}},\mathcal{A})$,
where the initial positions are specified
by the grounded spawn anchors.

Following the control-space diffusion
architecture of VBD~\cite{vbd},
we train a joint diffusion model on
the Waymo Open Motion Dataset (WOMD).
The diffusion process operates on
normalized control sequences.
Let $\mathcal{U}_0=
\operatorname{Norm}(\mathcal{A})$
denote the clean control sequence
and $\mathcal{U}_k$ its noisy counterpart
at diffusion step $k$.
The forward diffusion process is
\begin{equation}
\mathcal{U}_k
=
\sqrt{\bar{\alpha}_k}\mathcal{U}_0
+
\sqrt{1-\bar{\alpha}_k}\epsilon,
\quad
\epsilon\sim\mathcal{N}(0,\mathbf{I})
\label{eq:forward_diffusion}
\end{equation}
where $\alpha_k=1-\beta_k$,
$\bar{\alpha}_k=\prod_{j=1}^{k}\alpha_j$,
and $\beta_k$ is the predefined
noise variance.

A conditional denoiser predicts the
clean control sequence:
\begin{equation}
\hat{\mathcal{U}}_0
=
G_{\vartheta}(\mathcal{U}_k,k,c)
\label{eq:denoiser}
\end{equation}
where $\vartheta$ denotes the network
parameters and $c$ represents the encoded
motion-history and scene features.
The grounded routes and spatial anchors
provide additional guidance during inference.

The predicted controls are denormalized
and reconstructed into vehicle trajectories
through differentiable dynamics:
\begin{equation}
\hat{\mathcal{X}}_0
=
\operatorname{Rollout}
\left(
\mathcal{X}_{\mathrm{init}},
\operatorname{Denorm}(\hat{\mathcal{U}}_0)
\right)
\label{eq:predicted_rollout}
\end{equation}

The denoiser is trained by minimizing
the trajectory reconstruction loss:
\begin{equation}
\mathcal{L}_{\mathrm{train}}
=
\mathbb{E}_{\mathcal{S},k,\epsilon}
\left[
\operatorname{SmoothL1}
\left(
\hat{\mathcal{X}}_0-
\mathcal{X}^{\mathrm{gt}}
\right)
\right]
\label{eq:training_loss}
\end{equation}
where $\mathcal{S}$ denotes a training
scenario, $\mathcal{X}^{\mathrm{gt}}$
is the ground-truth trajectory,
and $\hat{\mathcal{X}}_0$ is the
trajectory reconstructed from
the predicted control sequence.
The variables $k$ and $\epsilon$
denote the sampled diffusion step
and Gaussian noise, respectively.

During inference, gradient-based guidance
incorporates the grounded spatial constraints
and interaction semantics into reverse diffusion.
The guidance energy is defined as
\begin{equation}
E_{\mathrm{guide}}
=
E_{\mathrm{geo}}
+
E_{\mathrm{int}}
+
E_{\mathrm{spd}}
\label{eq:guidance_energy}
\end{equation}
where $E_{\mathrm{geo}}$,
$E_{\mathrm{int}}$, and $E_{\mathrm{spd}}$
denote geometric, interaction,
and speed consistency costs,
respectively.

\paragraph{Geometric Consistency}
The geometric guidance encourages
vehicles to follow their designated
routes and reach the target poses:
\begin{equation}
E_{\mathrm{geo}}
=
\lambda_r C_{\mathrm{route}}
+
\lambda_g C_{\mathrm{pose}}
\label{eq:geometric_guidance}
\end{equation}

The route consistency cost is defined as
\begin{equation}
C_{\mathrm{route}}
=
\sum_{i\in\mathcal{I}}
\frac{
\sum_t m_{i,t}
\varphi^2(|d_{i,t}|-d_{\mathrm{tol}})
}{
\max(1,\sum_t m_{i,t})
}
\label{eq:route_cost}
\end{equation}
where $d_{i,t}$ denotes the lateral deviation
from the reference route, $m_{i,t}$
masks out the conflict region to allow
flexible interaction maneuvers, and
$d_{\mathrm{tol}}$ is the allowed deviation.
The function
$\varphi(z)=\kappa\operatorname{softplus}(z/\kappa)$
provides a smooth penalty.

The target-pose cost jointly evaluates
position and heading errors at each
trajectory step:
\begin{equation}
c_{i,t}^{\mathrm{pose}}
=
H\left(\frac{e_{i,t}^{p}}{l_p}\right)
+
H\left(\frac{e_{i,t}^{\theta}}{l_\theta}\right)
\label{eq:pose_step_cost}
\end{equation}
where $H(\cdot)$ denotes the Huber loss,
$l_p$ and $l_\theta$ are normalization
scales, and the position and heading
errors are defined as
$e_{i,t}^{p}=
\|\mathbf{p}_{i,t}-\mathbf{p}_i^\star\|_2$
and
$e_{i,t}^{\theta}=
|\operatorname{wrap}(\theta_{i,t}
-\theta_i^\star)|$, respectively.

A temporal soft-min aggregates
the joint pose errors:
\begin{equation}
C_{\mathrm{pose}}
=
-\beta\sum_{i\in\mathcal{I}}
\log\left(
\frac{1}{T+1}
\sum_{t=0}^{T}
\exp\left(
-\frac{c_{i,t}^{\mathrm{pose}}}{\beta}
\right)
\right)
\label{eq:pose_cost}
\end{equation}
where $\beta$ controls the
soft-min approximation.
This formulation encourages position
and heading consistency at the same
trajectory step.

\paragraph{Interaction Consistency}

The interaction guidance encourages
the prescribed passing order $\gamma$
between the ego vehicle and BV,
without imposing fixed entry times.

A differentiable first-entry estimator
is employed to estimate their entry
times into the conflict region.

The signed distance and soft entry
indicator are defined as
\begin{equation}
\begin{aligned}
d_{i,t}
&=\|\mathbf{p}_{i,t}
-\mathbf{p}_{\mathrm{conf}}\|_2-R,\\
h_{i,t}
&=\sigma\left(\frac{d_{i,t-1}}{\epsilon_e}\right)
\sigma\left(-\frac{d_{i,t}}{\epsilon_e}\right)
\end{aligned}
\label{eq:entry_hazard}
\end{equation}
where $R$ is the expanded conflict-region
radius, $\sigma(\cdot)$ denotes the sigmoid
function, and $\epsilon_e$ controls
the transition smoothness.
The entry indicator applies for
$t\geq 1$, with
$h_{i,0}=\sigma(-d_{i,0}/\epsilon_e)$.

The expected first-entry time is
estimated as
\begin{equation}
\hat t_i
=
\frac{
\sum_{t=0}^{T}t\Delta t\,h_{i,t}
\prod_{j<t}(1-h_{i,j})
}{
\underbrace{
1-\prod_{t=0}^{T}(1-h_{i,t})
}_{P_i}
+\varepsilon
}
\label{eq:first_entry}
\end{equation}
where $P_i$ denotes the estimated
entry probability, $\Delta t$ is
the sampling interval, and
$\varepsilon$ ensures numerical stability.

The entry-time difference is
\begin{equation}
\Delta t_\gamma
=
\hat t_{i_{\mathrm{second}}}
-
\hat t_{i_{\mathrm{first}}}
\label{eq:entry_time_difference}
\end{equation}
where $\gamma$ specifies the passing
order, with BV-first for rushing
and ego-first for yielding.

The interaction consistency cost is
\begin{equation}
E_{\mathrm{int}}
=\alpha P_{\mathrm E}P_{\mathrm B}
[\Delta t_\gamma]_-^2
+\zeta\sum_{i\in\mathcal I}(1-P_i)^2
\label{eq:interaction_guidance}
\end{equation}
where $\mathcal I=\{\mathrm E,\mathrm B\}$,
$[x]_-=\max(0,-x)$, and
$\alpha$ and $\zeta$ are weighting
coefficients.
The two terms penalize passing-order
violations and missing conflict-region
entries, respectively.

\paragraph{Speed Consistency}
The speed guidance encourages vehicles
to maintain their reference speeds
while penalizing violations of the
maximum speed limit:
\begin{equation}
\begin{aligned}
E_{\mathrm{spd}}
=
\sum_{i\in\mathcal{I}}
\sum_{t=0}^{T}
\Big[
&\lambda_v
(v_{i,t}-v_{i,t}^{\mathrm{ref}})^2+
\lambda_{\max}
\varphi^2(v_{i,t}-v_{\max})
\Big]
\end{aligned}
\label{eq:speed_guidance}
\end{equation}
where $v_{i,t}^{\mathrm{ref}}$ denotes
the reference speed of vehicle $i$
at time $t$, and $v_{\max}$ is the
maximum allowed speed.
$\lambda_v$ and $\lambda_{\max}$
are the corresponding guidance weights.

The guidance energy is evaluated on
trajectories reconstructed through
differentiable vehicle dynamics.
At each reverse diffusion step,
the scheduler-derived variable is
initialized as
$\mu_k^{(0)}
=
\mathcal{Q}_k(\mathcal{U}_k,
\hat{\mathcal{U}}_0)$
and refined through
\begin{equation}
\begin{aligned}
\mathbf{g}_k^{(r)}
&=
\nabla_{\mu_k^{(r)}}
E_{\mathrm{guide}}
\big(\mathcal{X}(\mu_k^{(r)})\big),\\
\mu_k^{(r+1)}
&=
\operatorname{AdamStep}
\big(\mu_k^{(r)},\mathbf{g}_k^{(r)}\big)
\end{aligned}
\label{eq:guided_diffusion}
\end{equation}
where $\mathcal{Q}_k$ denotes the diffusion
scheduler update, and
$\mathcal{X}(\mu_k^{(r)})$ denotes the
trajectory reconstructed from the
denormalized control variable.
The Adam update minimizes the guidance
energy.
The resulting trajectories serve as
initial references for subsequent
online interaction refinement.


\subsection{Online Interaction Refinement}
\label{sec:3.3}

During proving-ground testing, variations
in VUT behavior and trajectory-tracking
errors may alter the intended interaction
timing.
We employ a game-based online interaction
refinement method to adapt the initial
reference trajectories to real-time VUT
responses, enabling steerable interaction
risk during proving-ground testing.

The interaction is modeled as a
Stackelberg game, where the BV acts as
the leader $L$ and the VUT is represented
by a surrogate follower $F$.
The diffusion-generated BV and ego
trajectories serve as the leader and
follower references, respectively,
after transformation into their
corresponding Frenet coordinate systems.

For vehicle $i\in\{L,F\}$, the state
and control vectors are defined as
$\mathbf{x}^i=
[s^i,\dot{s}^i,l^i,\dot{l}^i]^\top$
and
$\mathbf{u}^i=
[\ddot{s}^i,\ddot{l}^i]^\top$,
respectively.
The vehicle dynamics are approximated by
\begin{equation}
\mathbf{x}_{j+1}^i
=
A\mathbf{x}_{j}^i+
B\mathbf{u}_{j}^i
\label{eq:game_dynamics}
\end{equation}
where
\begin{equation}
A=
\begin{bmatrix}
1 & \Delta t_g & 0 & 0\\
0 & 1 & 0 & 0\\
0 & 0 & 1 & \Delta t_g\\
0 & 0 & 0 & 1
\end{bmatrix},
\quad
B=
\begin{bmatrix}
\frac{\Delta t_g^2}{2} & 0\\
\Delta t_g & 0\\
0 & \frac{\Delta t_g^2}{2}\\
0 & \Delta t_g
\end{bmatrix},
\label{eq:game_matrices}
\end{equation}
and $\Delta t_g$ denotes the game
planning interval.
Over a prediction horizon of $H$ steps,
the state and control sequences are
denoted by
$\mathbf{X}^i=
(\mathbf{x}_1^i,\ldots,\mathbf{x}_H^i)$
and
$\mathbf{U}^i=
(\mathbf{u}_0^i,\ldots,\mathbf{u}_{H-1}^i)$,
respectively.
The planning variables are
$\mathbf{Z}_i=(\mathbf{X}^i,\mathbf{U}^i)$,
with the initial state fixed to the
latest observation.

The leader balances adversarialness
and realism through
\begin{equation}
J_L
=
\omega_{\mathrm{adv}}J_{\mathrm{adv}}
+
\omega_{\mathrm{real}}J_{\mathrm{real}}
\label{eq:leader_objective}
\end{equation}
where $\omega_{\mathrm{adv}}$ and
$\omega_{\mathrm{real}}$ denote the
corresponding weighting coefficients.

\paragraph{Adversarialness}
Following the anticipated post-encroachment
time (APET)~\cite{zhang2025testing},
we estimate the signed arrival-time gap
at the conflict point as
\begin{equation}
\Delta T_{\mathrm{conf},j}
=
\frac{s_{\mathrm{conf}}^F-s_{j+1}^F}
{\max(\dot{s}_{j+1}^F,\epsilon)}
-
\frac{s_{\mathrm{conf}}^L-s_{j+1}^L}
{\max(\dot{s}_{j+1}^L,\epsilon)}
\label{eq:apet}
\end{equation}
where $s_{\mathrm{conf}}^i$ denotes
the conflict-point station along
vehicle $i$'s reference route,
and $\epsilon$ is a small positive
constant.
A positive gap indicates that the
BV is predicted to arrive first,
whereas a negative gap indicates
that the VUT is predicted to
arrive first.

The adversarial cost penalizes
deviations from the prescribed
interaction timing:
\begin{equation}
J_{\mathrm{adv}}
=
\sum_{j=0}^{H-1}
\left(
\Delta T_{\mathrm{conf},j}
-
\sigma_{\gamma}\tau
\right)^2
\label{eq:directional_gap}
\end{equation}
where $\tau$ denotes the target APET
and $\sigma_{\gamma}=+1$ and $-1$
correspond to BV-first and VUT-first
passing orders, respectively.

\paragraph{Realism}
The realism cost encourages consistency
with the diffusion-generated reference
trajectory while penalizing control effort:
\begin{equation}
J_{\mathrm{real}}
=
\sum_{j=0}^{H-1}
\left(
\|\mathbf{x}_{j+1}^{L}
-\mathbf{x}_{\mathrm{ref},j+1}^{L}\|_{Q_L}^2
+
\|\mathbf{u}_{j}^{L}\|_{R_L}^2
\right)
\label{eq:realism}
\end{equation}
where $Q_L$ and $R_L$ are weighting matrices.

The follower predicts the VUT response
by minimizing deviations from its
reference trajectory and control effort:
\begin{equation}
\begin{aligned}
\min_{\mathbf{Z}_F}\quad
&J_F=
\sum_{j=0}^{H-1}
\Big(
\|\mathbf{x}_{j+1}^{F}
-\mathbf{x}_{\mathrm{ref},j+1}^{F}\|_{Q_F}^2\\
&\qquad+
\|\mathbf{u}_{j}^{F}\|_{R_F}^2
\Big)\\
\text{s.t.}\quad
&\mathbf{x}_{j+1}^{F}
=A\mathbf{x}_{j}^{F}+B\mathbf{u}_{j}^{F},\\
&\mathbf{u}_{\min}^{F}
\leq\mathbf{u}_{j}^{F}
\leq\mathbf{u}_{\max}^{F},\\
&\mathbf{n}_{j+1}^{\top}
\left(
\begin{bmatrix}
\tilde{s}_{j+1}^{F\to L}\\
\tilde{l}_{j+1}^{F\to L}
\end{bmatrix}
-
\begin{bmatrix}
s_{j+1}^{L}\\
l_{j+1}^{L}
\end{bmatrix}
\right)
\geq d_{\mathrm{safe},j+1},\\
&j=0,\ldots,H-1.
\end{aligned}
\label{eq:follower_opt}
\end{equation}
where $Q_F$ and $R_F$ are positive
semidefinite weighting matrices,
and $\mathbf{u}_{\min}^{F}$ and
$\mathbf{u}_{\max}^{F}$ denote
the acceleration bounds.
The follower position is projected
onto the leader's Frenet frame,
while $\mathbf{n}_j$ and
$d_{\mathrm{safe},j}$ denote the
fixed separation direction and
safety margin, respectively.
With a fixed affine projection,
the follower optimization forms
a convex quadratic program,
whose equality and inequality
constraints are denoted by
$h_F=0$ and $\mathbf{g}_F\geq0$.

As the VUT is a black-box ADS,
the follower serves as a surrogate
response model rather than a
directly controlled vehicle.
The predicted follower response
is incorporated into the leader
optimization through the
Karush--Kuhn--Tucker (KKT) conditions.
The resulting single-level
optimization problem is formulated as
\begin{equation}
\begin{aligned}
\min_{\mathbf{Z}_L,\mathbf{Z}_F,
\boldsymbol{\lambda},\boldsymbol{\mu}}
\quad
&J_L(\mathbf{Z}_L,\mathbf{Z}_F)\\
\text{s.t.}\quad
&h_L=0,\quad \mathbf{g}_L\geq0,\\
&h_F=0,\quad \mathbf{g}_F\geq0,\\
&\nabla_{\mathbf{Z}_F}
\bigl(
J_F+\boldsymbol{\lambda}^{\top}h_F
-\boldsymbol{\mu}^{\top}\mathbf{g}_F
\bigr)=0,\\
&\boldsymbol{\mu}\geq0,\\
&\boldsymbol{\mu}\odot\mathbf{g}_F=0.
\end{aligned}
\label{eq:leader_opt}
\end{equation}
where $h_L$ and $\mathbf{g}_L$
denote the leader's dynamics and
feasibility constraints.
The vectors $\boldsymbol{\lambda}$
and $\boldsymbol{\mu}$ are the
Lagrange multipliers associated
with the follower's equality and
inequality constraints, respectively,
and $\odot$ denotes element-wise
multiplication.

The optimization problem is formulated
in CasADi and solved using IPOPT.
As summarized in Algorithm~\ref{alg:testing},
the BV trajectory is refined at each
planning cycle using the latest
observed vehicle states.

\begin{algorithm}[t]
\caption{Overall Workflow of Video2Track}
\label{alg:testing}

\KwIn{
Scenario semantics $(\Psi,e,b,\phi,\gamma)$,
topology library $\mathcal{D}$,
target APET $\tau$,
initial states $\mathcal{X}_{\mathrm{init}}$,
diffusion condition $c$,
diffusion steps $K$,
guidance iterations $R_d$
}

\KwOut{
Initial reference trajectories
$\mathbf{X}_{\mathrm{ref}}^F,
\mathbf{X}_{\mathrm{ref}}^L$
and online refined BV trajectory
$\mathbf{X}_{L,\mathrm{UTM}}^*$
}

$\mathcal{M}^*
\leftarrow
\operatorname{Retrieve}
(\Psi,e,b,\phi;\mathcal{D})$\;

$\mathcal{C}_{\mathrm{map}}
\leftarrow
\operatorname{Ground}(\mathcal{M}^*)$\;

$\mathcal{U}_K
\sim\mathcal{N}(0,\mathbf{I})$\;

\For{$k=K,K-1,\ldots,1$}{

    $\hat{\mathcal{U}}_0
    \leftarrow
    G_{\vartheta}(\mathcal{U}_k,k,c)$\;

    $\mu_k^{(0)}
    \leftarrow
    \mathcal{Q}_k
    (\mathcal{U}_k,\hat{\mathcal{U}}_0)$\;

    \For{$r=0,\ldots,R_d-1$}{
    
        $\hat{\mathcal{X}}
        \leftarrow
        \operatorname{Rollout}
        \left(
        \mathcal{X}_{\mathrm{init}},
        \operatorname{Denorm}(\mu_k^{(r)})
        \right)$\;
    
        $\mathbf{g}_k^{(r)}
        \leftarrow
        \nabla_{\mu_k^{(r)}}
        E_{\mathrm{guide}}
        (\hat{\mathcal{X}};
        \mathcal{C}_{\mathrm{map}},\gamma)$\;
    
        $\mu_k^{(r+1)}
        \leftarrow
        \operatorname{AdamStep}
        (\mu_k^{(r)},\mathbf{g}_k^{(r)})$\;
    }

    $\mathcal{U}_{k-1}
    \leftarrow\mu_k^{(R_d)}$\;
}

$(\mathbf{X}_{\mathrm{ref}}^F,\mathbf{X}_{\mathrm{ref}}^L)
\leftarrow
\operatorname{GenerateReferences}(\mathcal{U}_0)$\;

\While{testing episode active}{

    $(\mathbf{x}_0^F,\mathbf{x}_0^L)
    \leftarrow\operatorname{Observe}()$\;

    $J_{\mathrm{adv}}\leftarrow
    \displaystyle\sum_{j=0}^{H-1}
    \left(
    \Delta T_{\mathrm{conf},j}
    -\sigma_\gamma\tau
    \right)^2$\;

    $J_{\mathrm{real}}\leftarrow
    \displaystyle\sum_{j=0}^{H-1}
    \left(
    \|\mathbf{x}_{j+1}^{L}-
    \mathbf{x}_{\mathrm{ref},j+1}^{L}\|_{Q_L}^2
    +
    \|\mathbf{u}_j^L\|_{R_L}^2
    \right)$\;

    $J_L\leftarrow
    \omega_{\mathrm{adv}}J_{\mathrm{adv}}
    +\omega_{\mathrm{real}}J_{\mathrm{real}}$\;

    $J_F\leftarrow
    \displaystyle\sum_{j=0}^{H-1}
    \left(
    \|\mathbf{x}_{j+1}^{F}-
    \mathbf{x}_{\mathrm{ref},j+1}^{F}\|_{Q_F}^2
    +
    \|\mathbf{u}_j^F\|_{R_F}^2
    \right)$\;

    $(\mathbf{Z}_L^*,\mathbf{Z}_F^*)
    \leftarrow
    \operatorname{SolveKKT}
    (J_L,J_F;
    h_L,\mathbf{g}_L,h_F,\mathbf{g}_F)$\;

    $\mathbf{X}_{L,\mathrm{UTM}}^*
    \leftarrow
    \operatorname{FrenetToUTM}
    \left(\{\mathbf{x}_j^{L,*}\}_{j=1}^{H}\right)$\;

    $\operatorname{MQTT.Send}
    (\mathbf{X}_{L,\mathrm{UTM}}^*)$\;

}

\end{algorithm}

\begin{figure}
	\centering
	\includegraphics[width=\columnwidth]{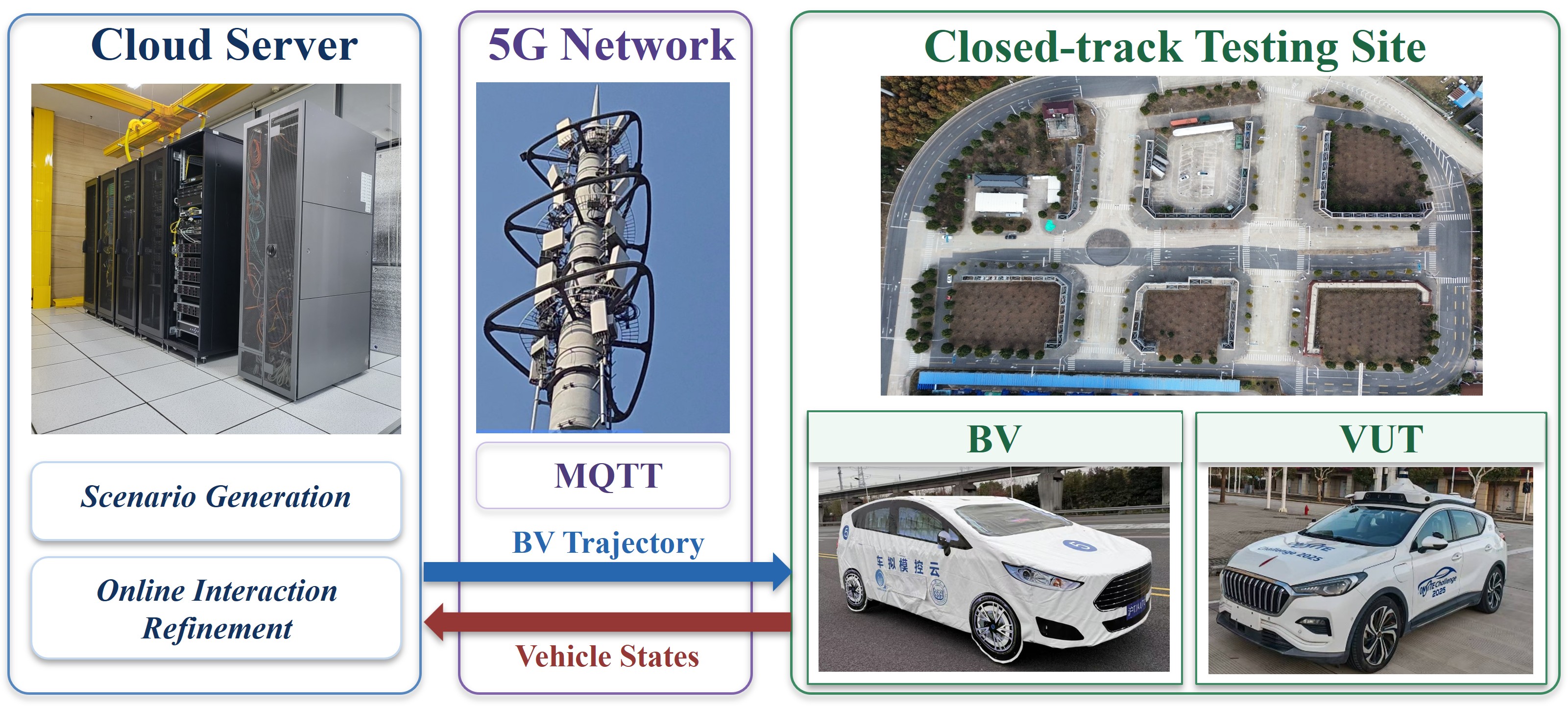}
	\caption{Deployment of Video2Track on the cloud-controlled proving-ground testing platform}
	\label{FIG:implementation}
\end{figure}

\begin{figure}[!t]
    \centering

    \includegraphics[
        width=\columnwidth,
        trim=0 3 0 0,
        clip
    ]{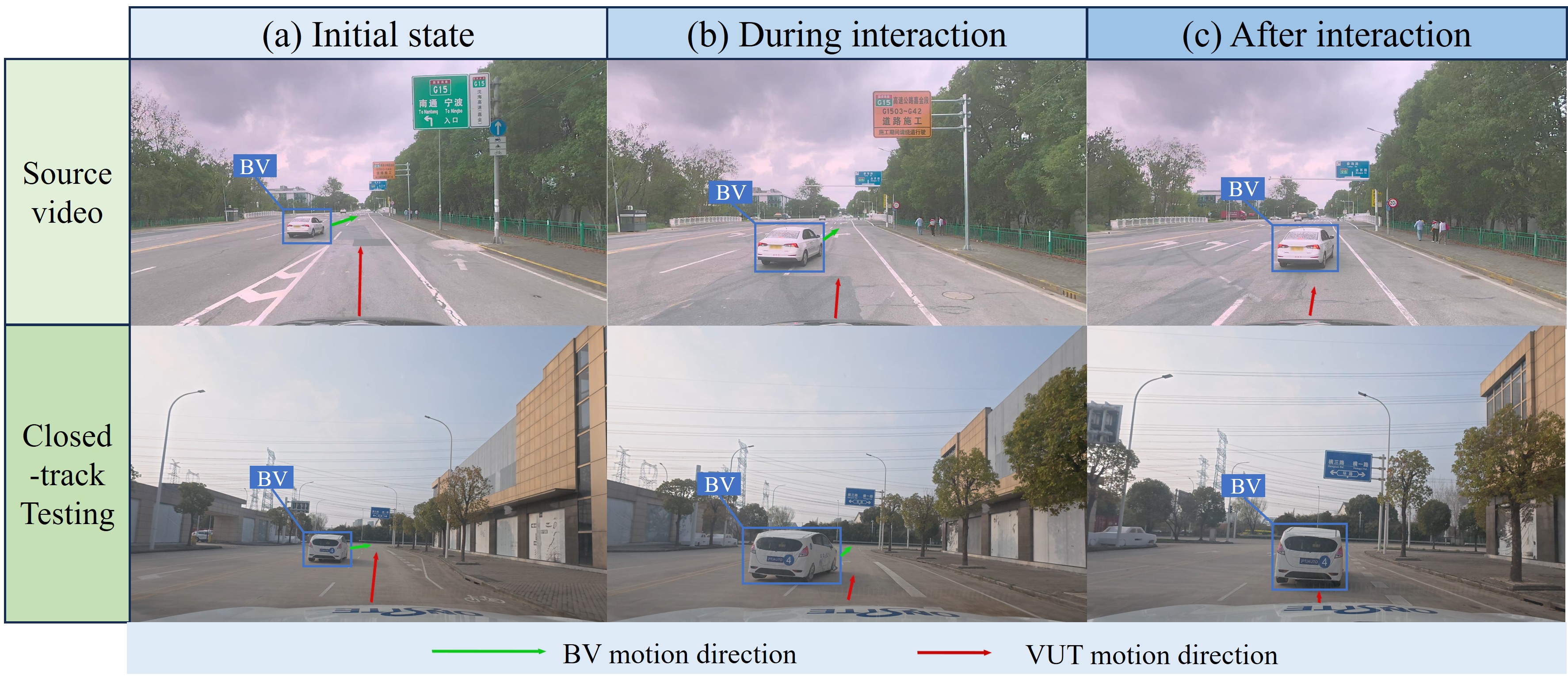}

    \caption{Comparison between the source video
    and proving-ground execution in the
    cut-in scenario.}
    \label{FIG:cut-in-photo}

    \vspace{0.6em}

    \subfloat[]{%
        \includegraphics[width=0.48\columnwidth]
        {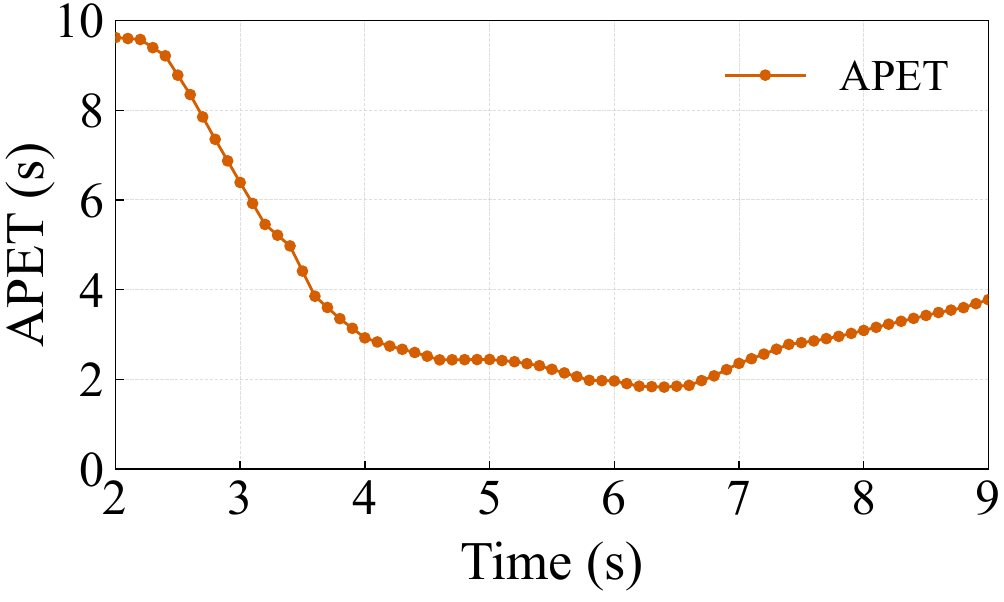}%
        \label{FIG:merge_APET_closed_track}%
    }%
    \hfill
    \subfloat[]{%
        \includegraphics[width=0.48\columnwidth]
        {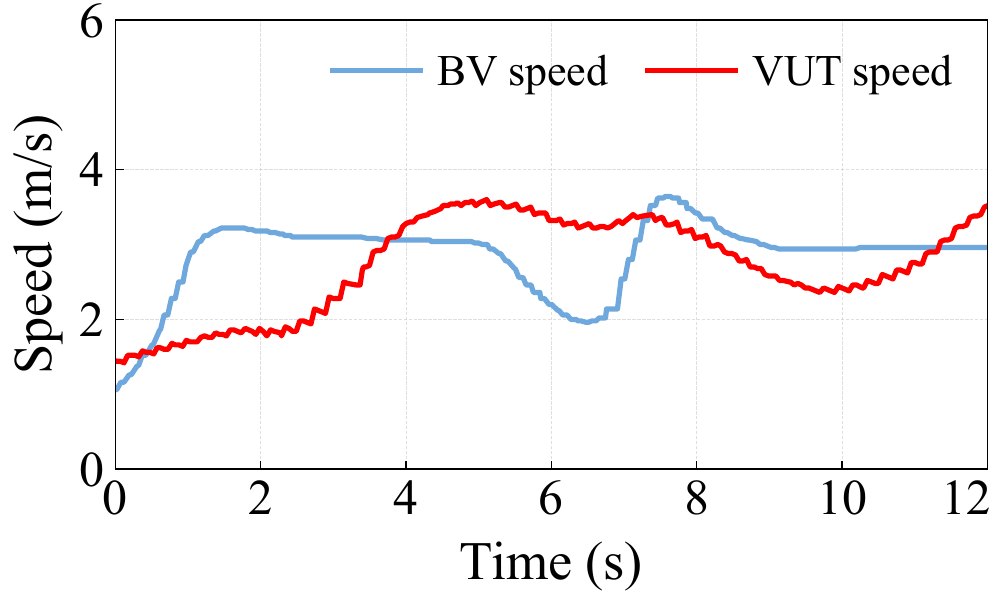}%
        \label{FIG:merge_Speed_closed_track}%
    }%

    \par\vspace{0.05em}

    \subfloat[]{%
        \includegraphics[width=0.95\columnwidth]
        {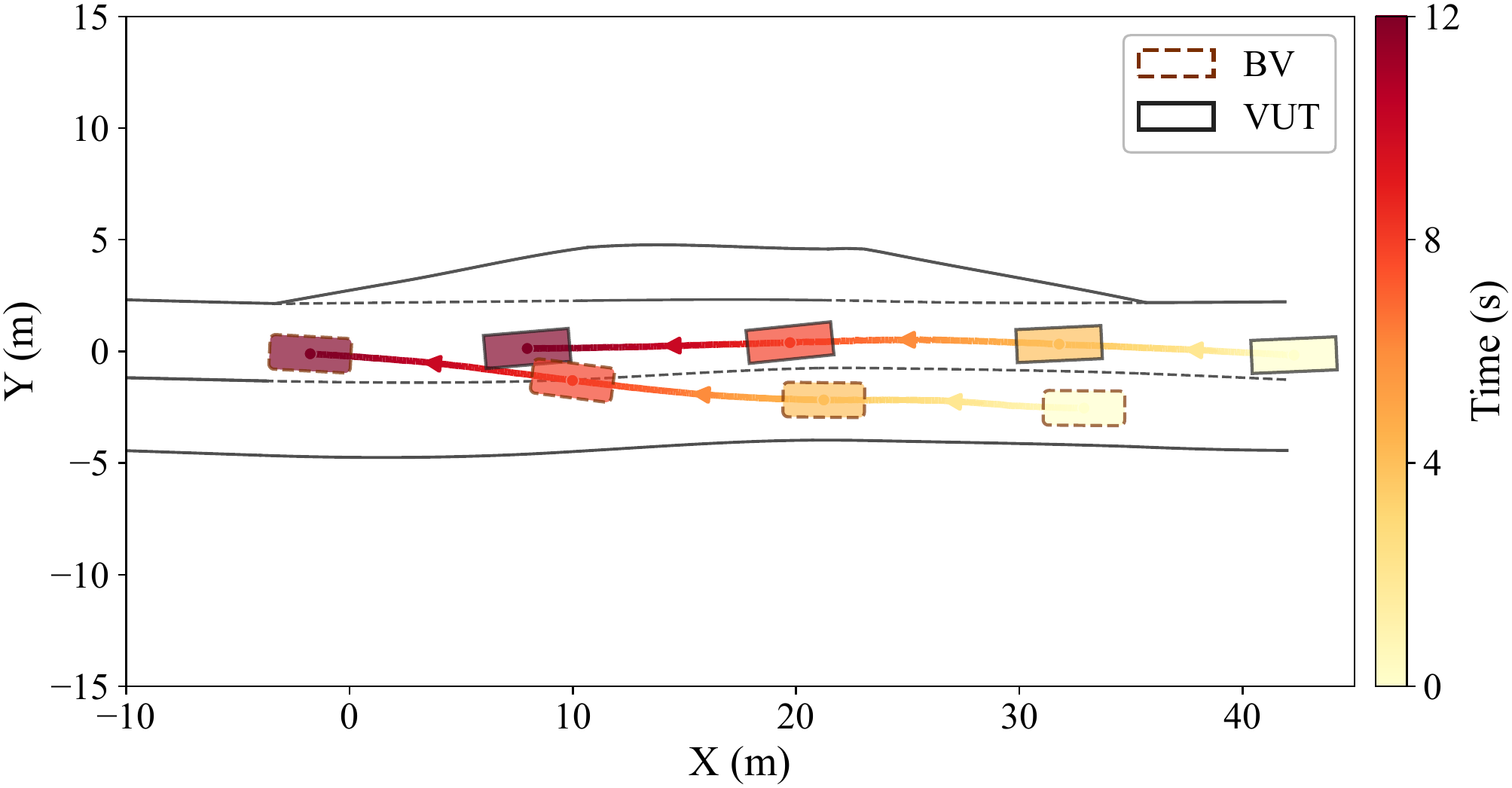}%
        \label{FIG:merge_Traj_closed_track}%
    }%

    \caption{Proving-ground execution results
    for the cut-in scenario:
    (a) APET evolution;
    (b) vehicle speed profiles;
    (c) vehicle trajectories.}
    \label{FIG:merge_PET_closed_track}
\end{figure}

\section{Experiments}\label{sec:4}

To evaluate Video2Track in terms of
fidelity, steerability, and adversariality,
we investigate the following research questions:

\begin{itemize}
    \item \textbf{RQ1:} Can Video2Track
    faithfully reproduce real-world driving
    interactions on a physical proving ground?

    \item \textbf{RQ2:} Can Video2Track
    generate scenarios with steerable
    interaction risk?

    \item \textbf{RQ3:} Can Video2Track
    effectively generate adversarial scenarios
    for stress testing?
\end{itemize}

\subsection{Experimental Setup}\label{sec:4.1}

\subsubsection{Platform and Deployment}
\label{sec:4.1.1}

The proposed Video2Track framework is deployed
at the Tongji University Intelligent Connected
Vehicle Testing and Evaluation Base, as shown
in Fig.~\ref{FIG:implementation}.
The cloud server, equipped with an
AMD EPYC 7763 CPU and an NVIDIA RTX 4090 GPU,
performs scenario generation and online
interaction refinement.
During proving-ground testing,
BV trajectories are refined online
based on VUT state feedback and transmitted
to the BV via a dedicated 5G network
using MQTT with QoS 1.
The average communication round-trip time
(RTT) is approximately 40~ms.

The BV is equipped with a high-precision
GNSS/INS navigation system and a
Raspberry Pi-based onboard controller.
The BV uploads localization data
to the cloud server at 50~Hz,
while the VUT employs a portable VBOX
positioning system to provide vehicle-state
feedback via MQTT.
The BV onboard controller operates at
100~Hz, employing longitudinal PID and
lateral Stanley controllers to track
the received trajectories.
The maximum path-tracking error during
proving-ground testing is approximately
0.15~m.

\subsubsection{Compared Methods}
\label{sec:4.1.2}

We compare Video2Track with three baseline
methods to evaluate scenario generation
and adversarial testing capabilities.

\textbf{Chat2Scenic:}
An iterative retrieval-augmented scenario
generation framework~\cite{chat2scenic}
that translates natural-language descriptions
into executable Scenic programs.
These programs specify road topology,
spatial relations, and vehicle behaviors,
enabling map-based sampling and scenario
instantiation in CARLA.

\textbf{Predefined Scenario:}
A conventional proving-ground testing
baseline in which the BV follows a
predefined speed profile after a preset
trigger, without adapting its behavior
to real-time VUT responses~\cite{queiroz2024driver}.

\textbf{TroubleMaker:}
A game-based adversarial testing
framework~\cite{zhang2025real}
that dynamically adjusts BV behavior
in response to VUT feedback during
physical proving-ground testing.
It employs predefined testing
configurations and reference routes.

\subsubsection{Metrics}
\label{sec:4.1.3}

We evaluate scenario generation,
interaction fidelity, risk steerability,
and adversarial testing capability
using the following metrics.

\textbf{Scenario Generation Success Rate (SGSR):}
Motivated by prior evaluations of
scenario generation accuracy~\cite{chat2scenic},
we assess scenario generation using
topology consistency rate (TCR),
interaction consistency rate (ICR),
and scenario generation success rate (SGSR).
TCR denotes the percentage of scenarios
whose mapped road configurations match
the topology semantics of the source videos.
ICR denotes the percentage of scenarios
whose generated trajectories preserve
the specified vehicle maneuvers and
interaction relationships, including
passing orders.
SGSR measures the proportion of scenarios
satisfying both criteria, defined as
\begin{equation}
\mathrm{SGSR}
=
\frac{N_{\mathrm{success}}}
{N_{\mathrm{total}}}
\times 100\%,
\label{eq:sgsr}
\end{equation}
where $N_{\mathrm{success}}$ denotes
the number of scenarios satisfying
both topology and interaction consistency,
and $N_{\mathrm{total}}$ is the total
number of evaluated scenarios.

\textbf{Minimum APET:}
We adopt anticipated post-encroachment
time (APET)~\cite{zhang2025testing}
to evaluate interaction risk.
APET is computed from recorded trajectories
using (24) during the conflict interaction phase.
For each execution, the minimum absolute APET
is recorded, and its mean is reported
over repeated trials.
APET is capped at 10~s for visualization
purposes only.
Interaction fidelity and risk steerability
are evaluated by deviations from the
source-video and target APET, respectively.
Smaller mean minimum APET values indicate
more temporally critical interactions.

\subsubsection{Experiment Design}
\label{sec:4.1.4}

The experiments consider two representative
interaction types, cut-in and unprotected
left-turn, derived from real-world
driving videos.

For RQ1, we evaluate scenario generation
capability and cross-map applicability.
We compare Video2Track with Chat2Scenic
on the CARLA Town03 map, which contains
diverse road topologies, including
intersections, roundabouts, and straight
road segments.
Both methods use the same video-derived
interaction descriptions and target road
network, and are evaluated using TCR,
ICR, and SGSR.
By constructing the topology library
on Town03, we further examine whether
the proposed interaction-aware retrieval
method can be applied beyond the
specific proving-ground map.
Interaction fidelity is subsequently
assessed by comparing the scenarios
reproduced on the physical proving
ground with their source videos.

For RQ2, we evaluate interaction risk
steerability in the unprotected left-turn
scenario using target APET values of
2, 3, 4, and 5~s.
Each configuration is executed five times
on the proving ground.

For RQ3, we evaluate adversarial testing
capability in both cut-in and unprotected
left-turn scenarios.
Video2Track is compared with the
Predefined Scenario and TroubleMaker,
with five proving-ground executions
per method and scenario.

\begin{figure}[!t]
    \centering

    \includegraphics[width=\columnwidth]
    {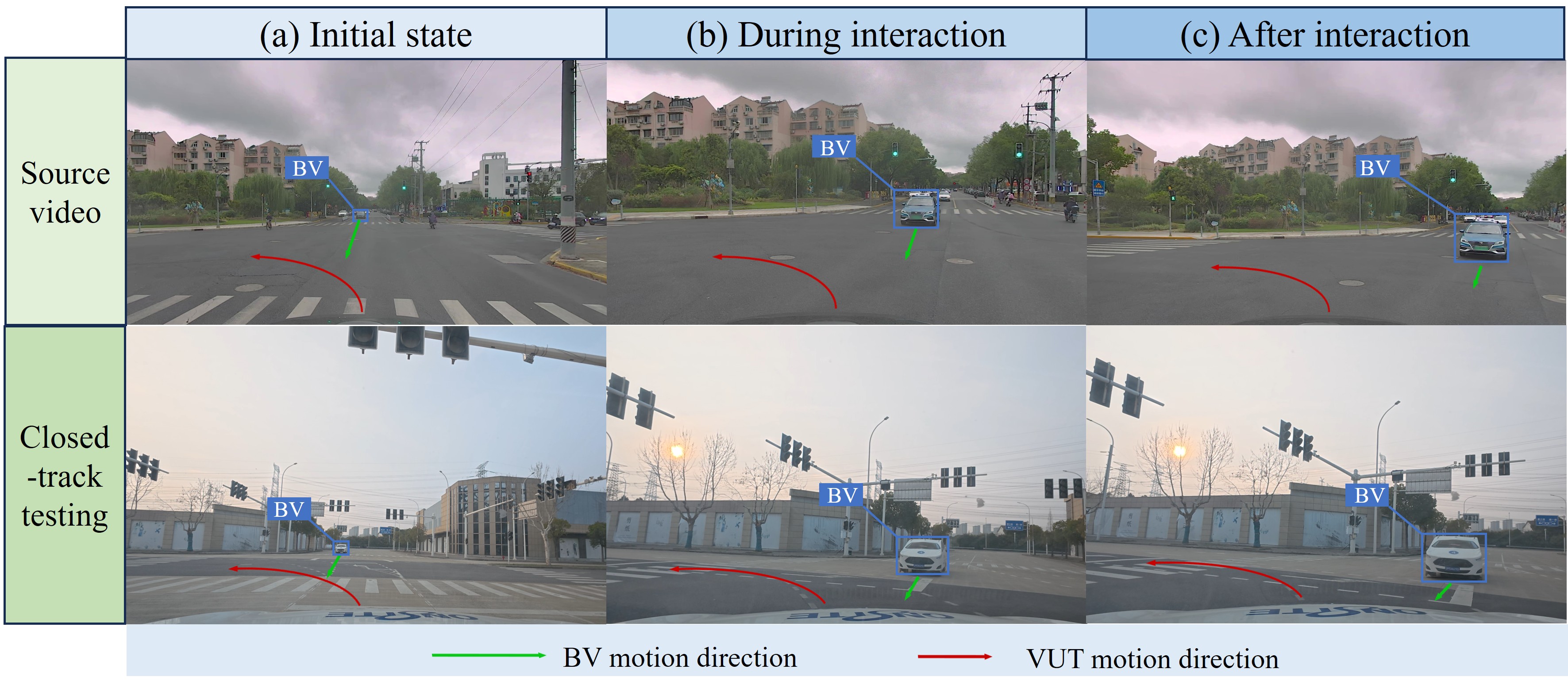}

    \caption{Comparison between the source video
    and proving-ground execution in the
    unprotected left-turn scenario.}
    \label{FIG:cross-photo}

    \vspace{0.6em}

    \subfloat[]{%
        \includegraphics[width=0.48\columnwidth]
        {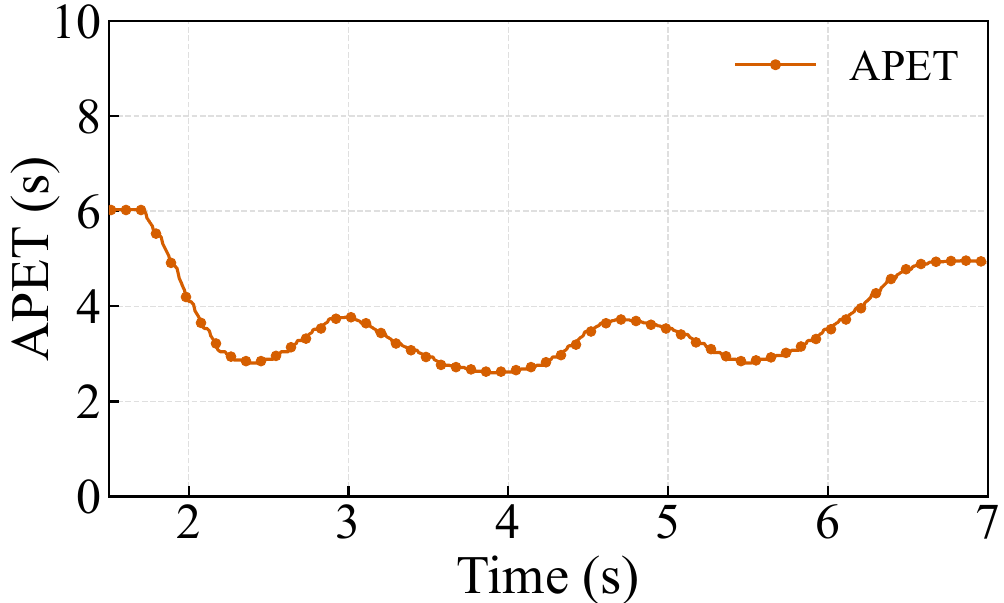}%
        \label{FIG:Cross_APET_closed_track}%
    }%
    \hfill
    \subfloat[]{%
        \includegraphics[width=0.48\columnwidth]
        {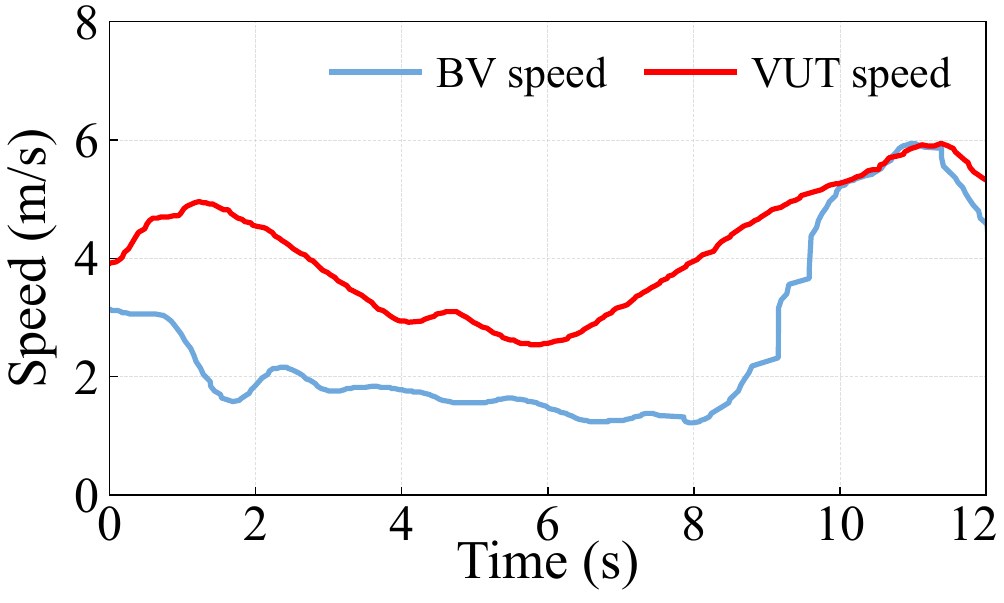}%
        \label{FIG:Cross_Speed_closed_track}%
    }%

    \par\vspace{0.05em}

    \subfloat[]{%
        \includegraphics[width=0.90\columnwidth]
        {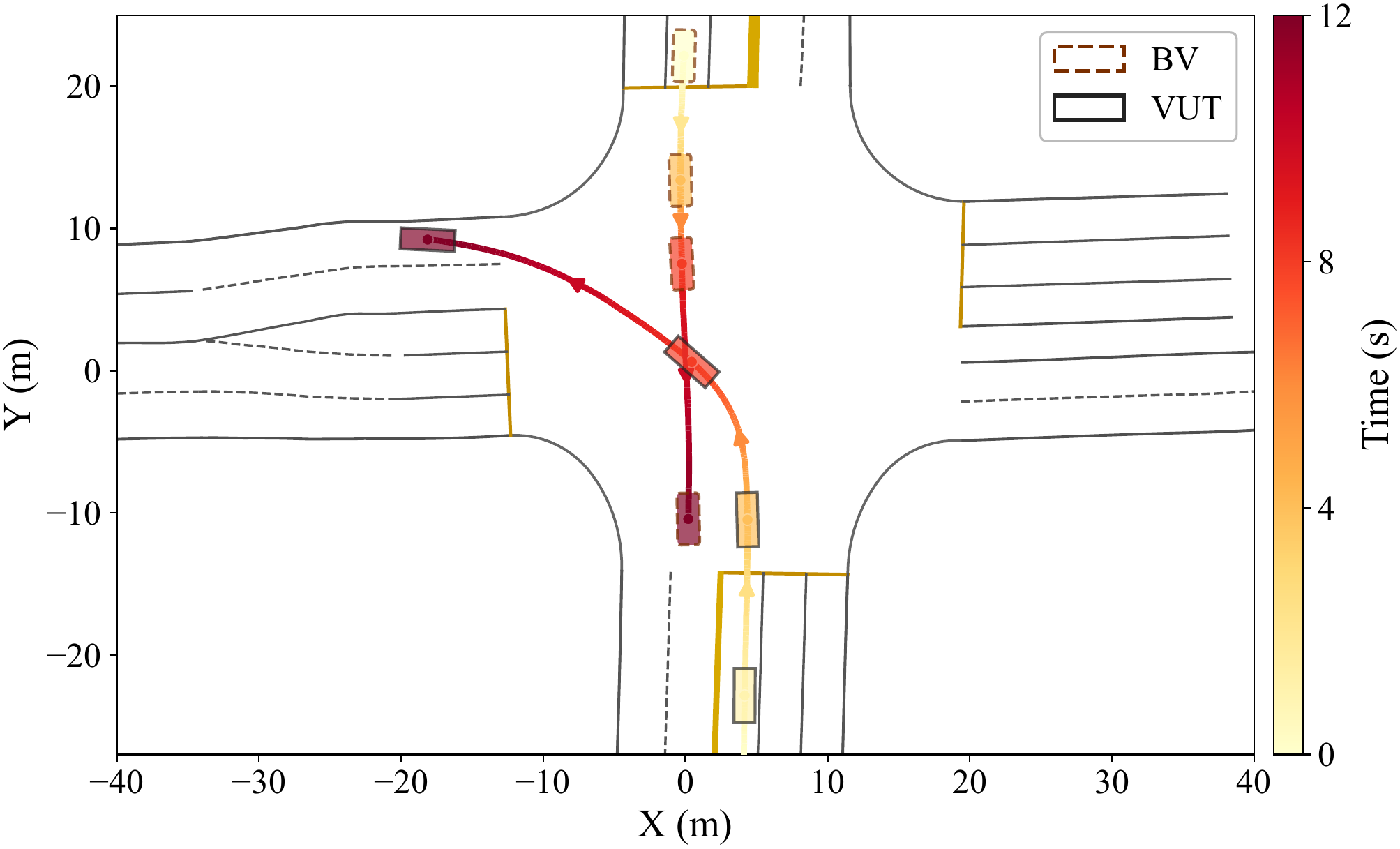}%
        \label{FIG:Cross_Traj_closed_track}%
    }%

    \caption{Proving-ground execution results
    for the unprotected left-turn scenario:
    (a) APET evolution;
    (b) vehicle speed profiles;
    (c) vehicle trajectories.}
    \label{FIG:Cross_PET_closed_track}
\end{figure}

\subsection{Results and Analysis}
\label{sec:4.2}

\subsubsection{RQ1: Interaction Reproduction Fidelity}
\label{sec:4.2.1}

\paragraph{Scenario Generation Success}

We compare Video2Track with Chat2Scenic
using identical video-derived interaction
descriptions and target road networks.
As shown in Table~\ref{tab:scenario_success},
Video2Track improves the TCR from 78.6\%
to 89.3\%, while the ICR increases
substantially from 50.0\% to 82.1\%.
Chat2Scenic relies on map-constrained
scenario sampling, which primarily ensures
topological compatibility but provides
limited control over vehicle maneuvers
and passing orders.
In contrast, Video2Track combines
interaction-aware maneuver-pair retrieval
with interaction-guided diffusion to
preserve the intended interaction
relationships.
This leads to an increase in SGSR
from 46.4\% to 78.6\%.

\begin{table}[t]
\centering
\caption{Comparison of scenario generation
success rates (\%).}
\label{tab:scenario_success}
\begin{tabular*}{\columnwidth}
{@{\extracolsep{\fill}}lccc}
\toprule
Method & TCR & ICR & SGSR \\
\midrule
Chat2Scenic & 78.6 & 50.0 & 46.4 \\
\textbf{Video2Track (Ours)}
& \textbf{89.3} & \textbf{82.1}
& \textbf{78.6} \\
\bottomrule
\end{tabular*}
\end{table}

\paragraph{Proving-Ground Reproduction}
We further evaluate interaction fidelity
by comparing representative frames
from the source videos and proving-ground
executions, as shown in
Figs.~\ref{FIG:cut-in-photo} and
\ref{FIG:cross-photo}.
In the cut-in scenario, the BV merges
from the adjacent lane while the VUT yields.
In the unprotected left-turn scenario,
the VUT turns left while the oncoming
BV yields.
Both scenarios preserve the vehicle
maneuvers and passing orders observed
in the source videos.

The corresponding trajectories,
speed profiles, and APET evolution
are presented in
Figs.~\ref{FIG:merge_PET_closed_track}
and \ref{FIG:Cross_PET_closed_track},
illustrating the reproduced interaction
dynamics during proving-ground testing.
The source-video APET values are provided
by the \textit{VideoScenario} dataset labels,
whereas the proving-ground APET values
are calculated from recorded vehicle trajectories.
In the cut-in scenario, the APET values
of the source video and proving-ground
execution are 1.90~s and 1.83~s,
respectively, yielding an absolute error
of 0.07~s.
For the unprotected left-turn scenario,
the corresponding values are 2.91~s
and 2.60~s, with an absolute error
of 0.31~s.
These results demonstrate that Video2Track
preserves the intended vehicle maneuvers
and passing orders while reproducing
interaction timing with APET errors
below 0.5~s in both evaluated scenarios.

\subsubsection{RQ2: Interaction Risk Steerability}
\label{sec:4.2.2}

\begin{figure}[!t]
    \centering

    \subfloat[Target APET = 2 s]{%
        \begin{minipage}[t]{\columnwidth}
            \centering
            \vspace{0pt}
            \includegraphics[
                width=0.48\linewidth,
                height=2.5cm,
                keepaspectratio
            ]{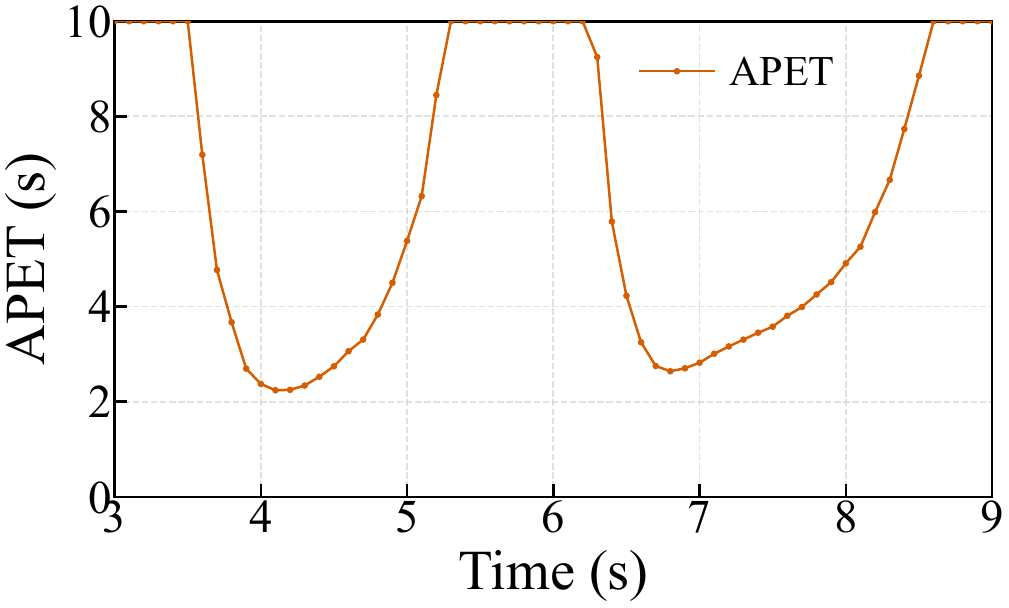}%
            \hfill
            \includegraphics[
                width=0.48\linewidth,
                height=2.5cm,
                keepaspectratio
            ]{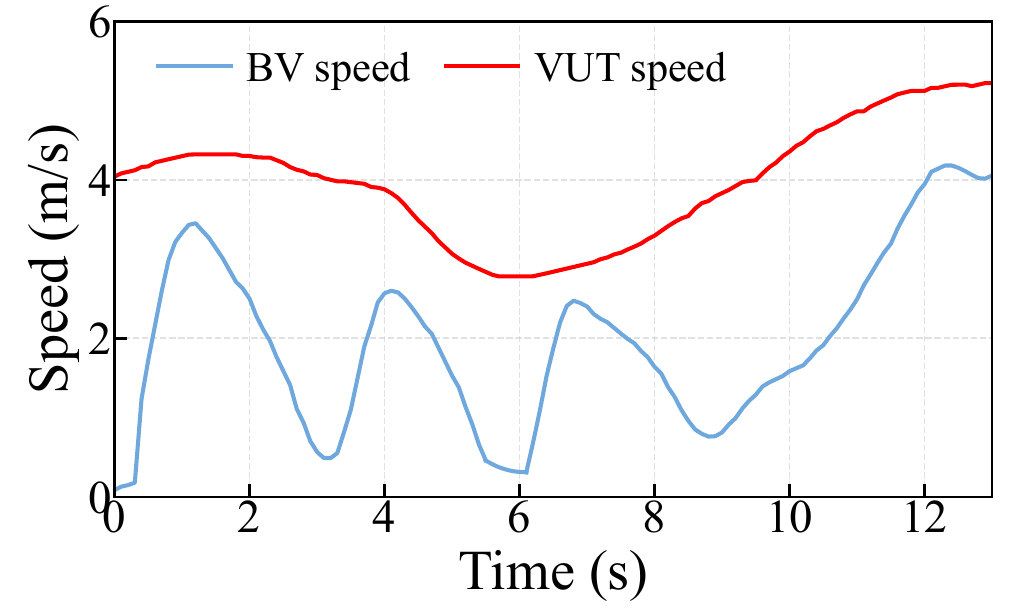}
        \end{minipage}%
        \label{fig:apet2}%
    }

    \par\vspace{0.2em}

    \subfloat[Target APET = 3 s]{%
        \begin{minipage}[t]{\columnwidth}
            \centering
            \vspace{0pt}
            \includegraphics[
                width=0.48\linewidth,
                height=2.5cm,
                keepaspectratio
            ]{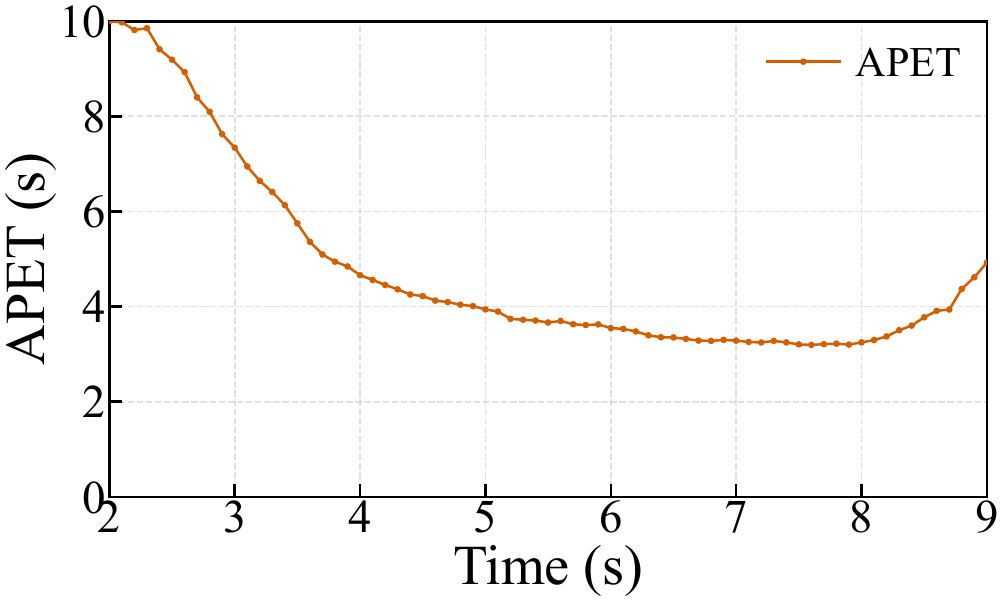}%
            \hfill
            \includegraphics[
                width=0.48\linewidth,
                height=2.5cm,
                keepaspectratio
            ]{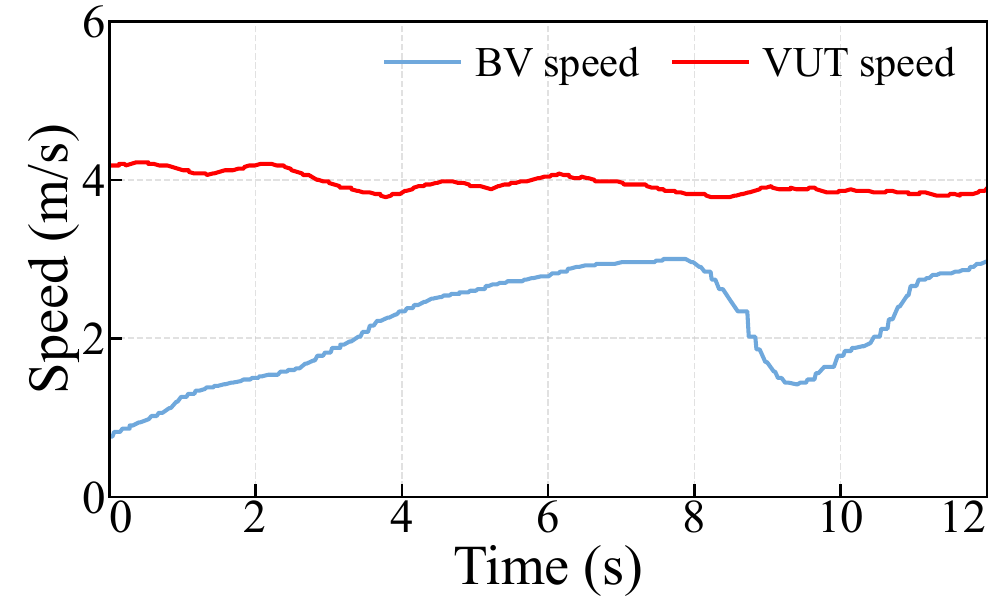}
        \end{minipage}%
        \label{fig:apet3}%
    }

    \par\vspace{0.2em}

    \subfloat[Target APET = 4 s]{%
        \begin{minipage}[t]{\columnwidth}
            \centering
            \vspace{0pt}
            \includegraphics[
                width=0.48\linewidth,
                height=2.5cm,
                keepaspectratio
            ]{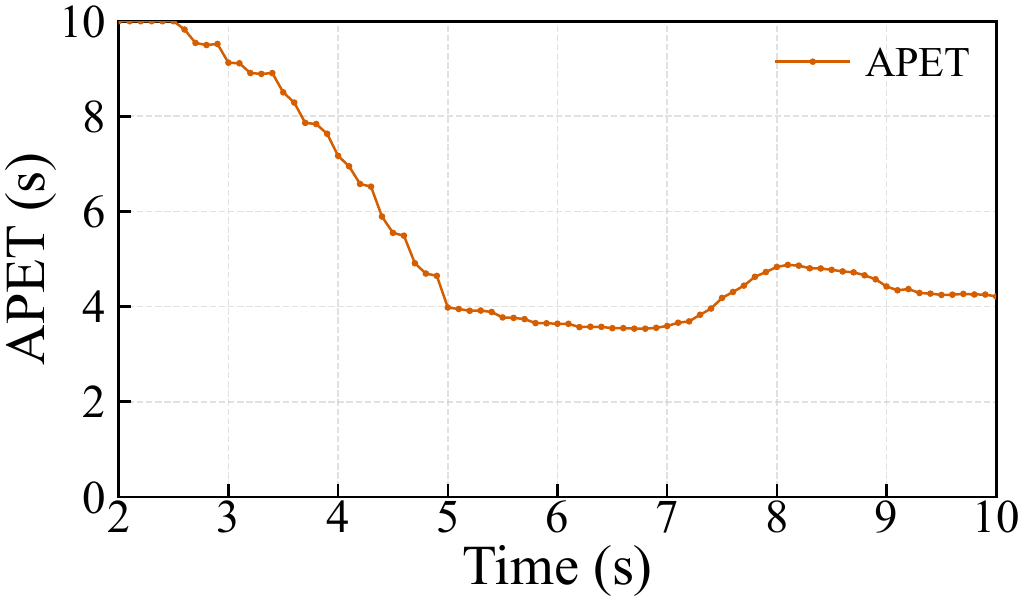}%
            \hfill
            \includegraphics[
                width=0.48\linewidth,
                height=2.5cm,
                keepaspectratio
            ]{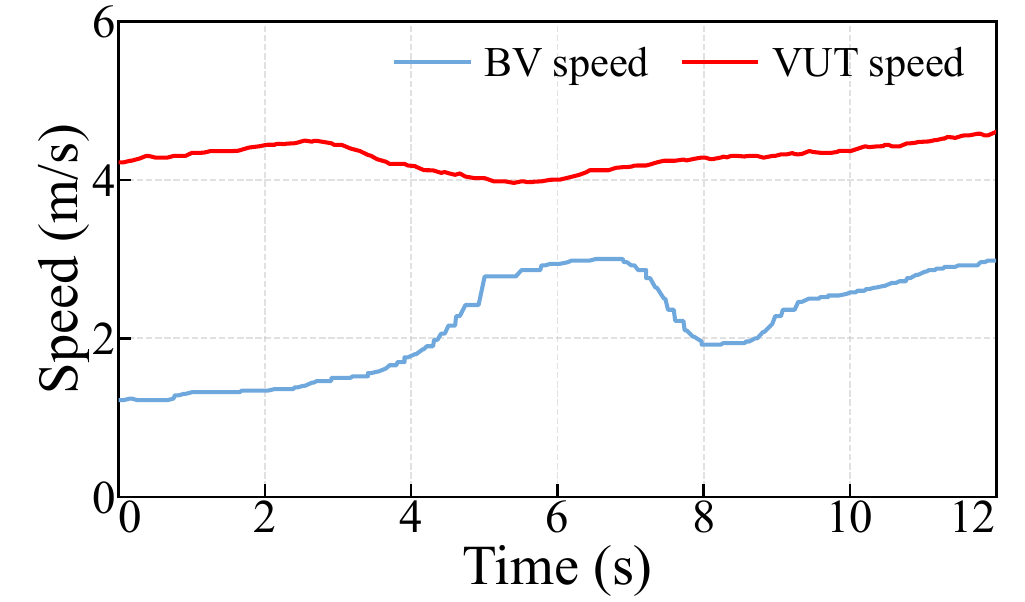}
        \end{minipage}%
        \label{fig:apet4}%
    }

    \par\vspace{0.2em}

    \subfloat[Target APET = 5 s]{%
        \begin{minipage}[t]{\columnwidth}
            \centering
            \vspace{0pt}
            \includegraphics[
                width=0.48\linewidth,
                height=2.5cm,
                keepaspectratio
            ]{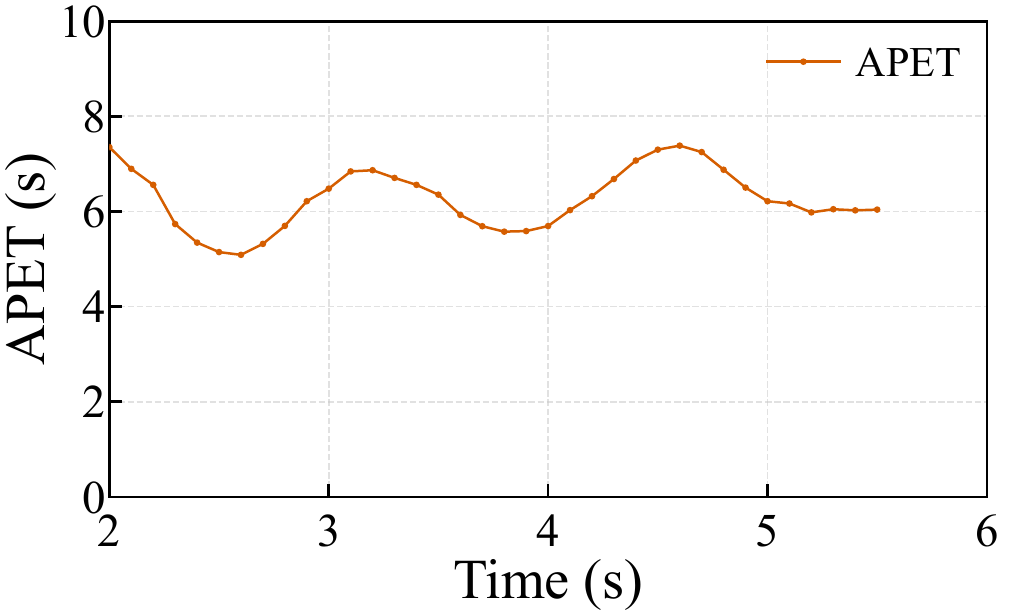}%
            \hfill
            \includegraphics[
                width=0.48\linewidth,
                height=2.5cm,
                keepaspectratio
            ]{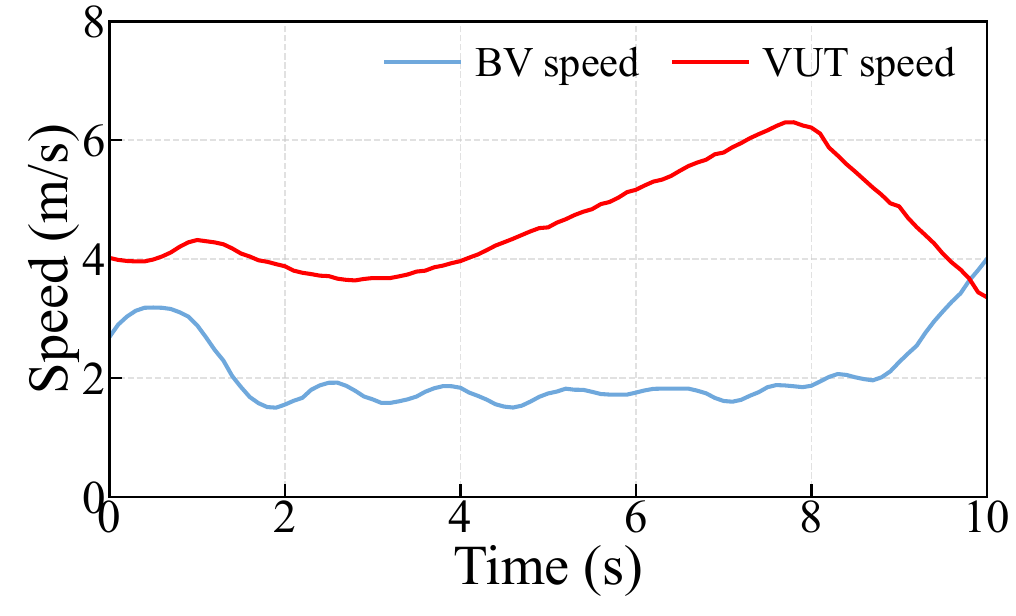}
        \end{minipage}%
        \label{fig:apet5}%
    }

    \caption{Interaction risk regulation
    under different target APET values
    during proving-ground testing.
    Panels (a)--(d) correspond to
    target APET values of 2, 3, 4,
    and 5~s, respectively.
    Each panel shows the APET evolution
    (left) and vehicle speed profiles
    (right).}
    \label{fig:four_images}
\end{figure}

\begin{figure}[!t]
    \centering
    \vspace{-0.32cm}
    \subfloat[APET]{%
        \includegraphics[
            width=0.48\columnwidth,
            height=2.5cm,
            keepaspectratio
        ]{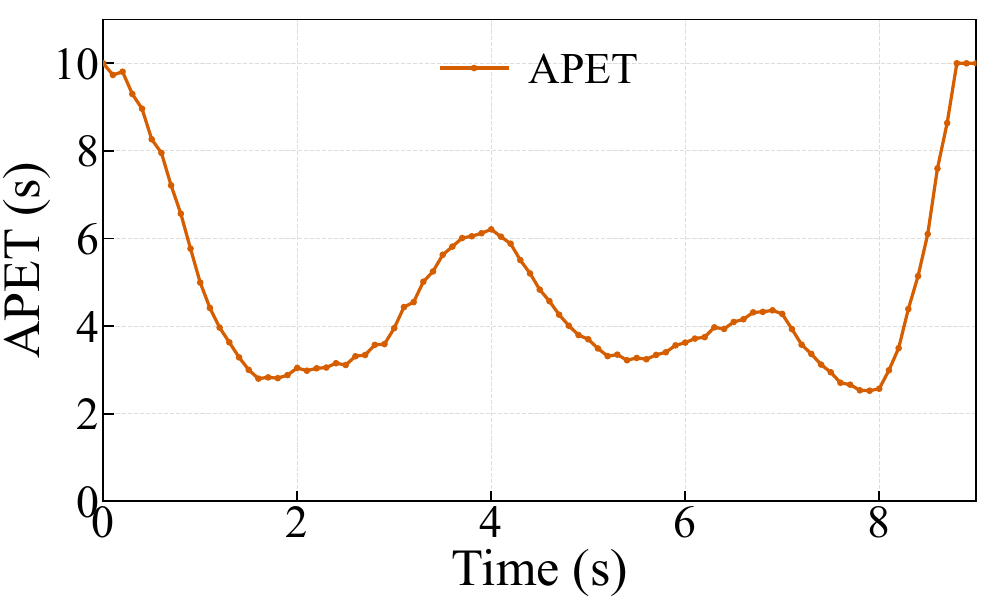}%
        \label{fig:rush_apet}%
    }%
    \hfill
    \subfloat[Speed]{%
        \includegraphics[
            width=0.48\columnwidth,
            height=2.5cm,
            keepaspectratio
        ]{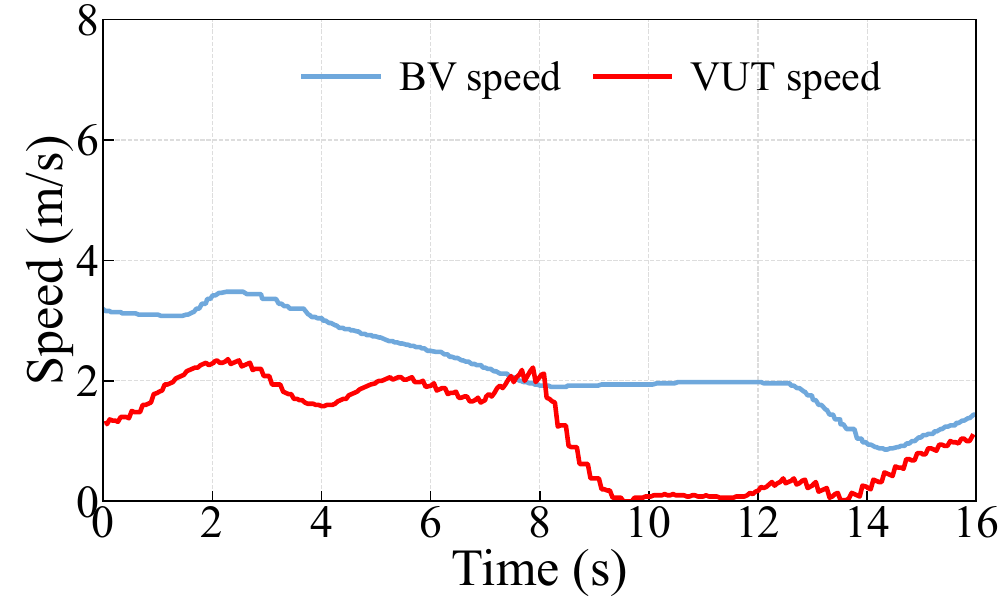}%
        \label{fig:rush_speed}%
    }

    \par\vspace{0.2em}

    \subfloat[Trajectories]{%
        \includegraphics[
            width=0.95\columnwidth
        ]{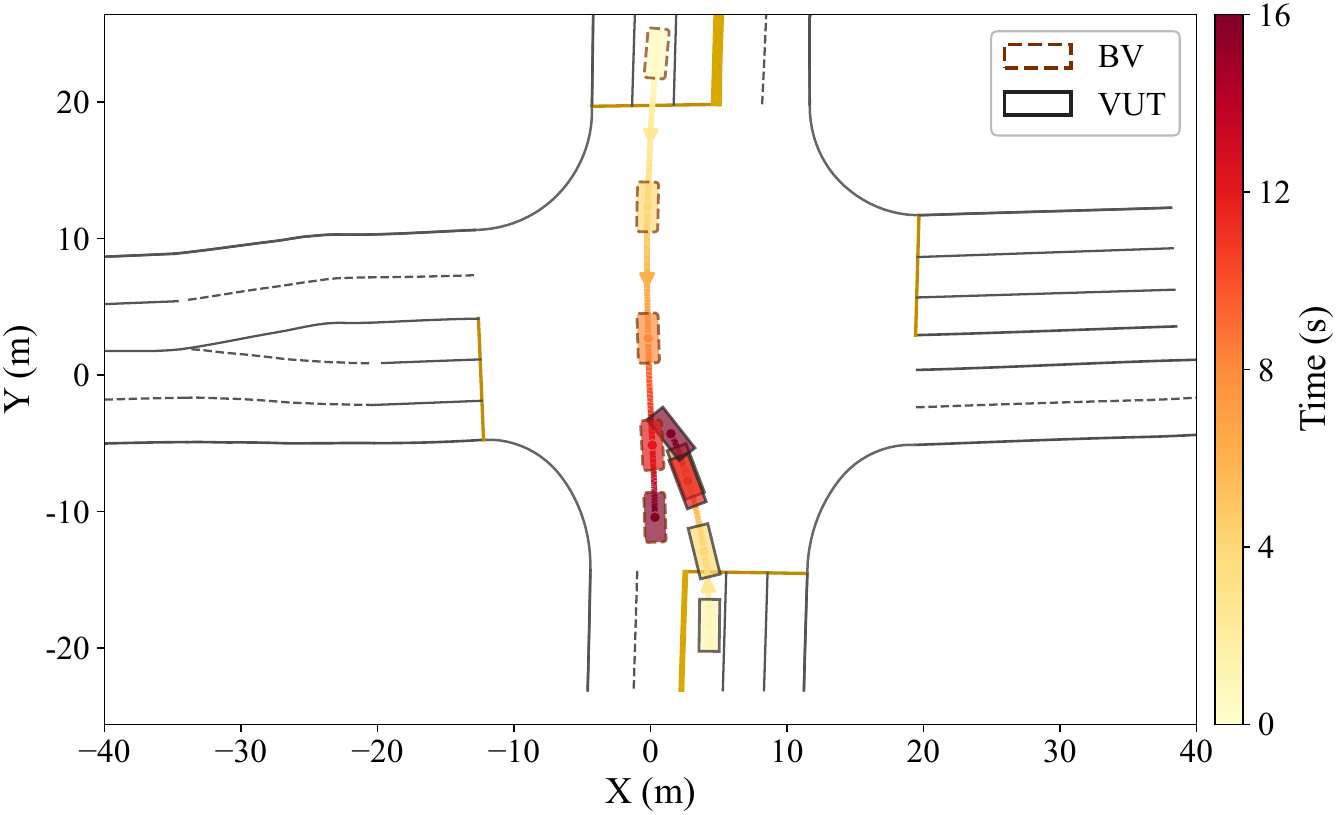}%
        \label{fig:rush_trajectory}%
    }

    \caption{Proving-ground execution of
    the rushing variant in the
    unprotected left-turn scenario.
    (a) APET evolution.
    (b) Vehicle speed profiles.
    (c) Vehicle trajectories.}
    \label{fig:cross_rush}
\end{figure}

\begin{table}[t]
\centering
\caption{APET regulation results in the unprotected
left-turn scenario.}
\label{tab:apet_steerability}
\begin{tabular*}{\columnwidth}
{@{\extracolsep{\fill}}ccc@{}}
\toprule
Target (s) & Mean Minimum APET (s) & Absolute Error (s) \\
\midrule
2.0 & 2.24 & 0.24 \\
3.0 & 3.19 & 0.19 \\
4.0 & 3.54 & 0.46 \\
5.0 & 5.09 & 0.09 \\
\bottomrule
\end{tabular*}
\end{table}

We evaluated interaction risk steerability in the
unprotected left-turn scenario at target APET values
of 2, 3, 4, and 5~s, with five proving-ground trials
per target. Table~\ref{tab:apet_steerability}
reports the mean minimum APET across five trials
at each target and the absolute error between
this mean and the prescribed target.
The mean minimum APET increased from
2.24 to 5.09~s as the target increased
from 2 to 5~s. The absolute error ranged
from 0.09 to 0.46~s across the four
target settings.

Fig.~\ref{fig:four_images} shows the APET and
vehicle speed profiles from one representative trial
at each target. In the trial with a 2.0~s target,
the BV accelerated and decelerated as it approached
the conflict region, while the VUT slowed during
the interaction. These profiles illustrate how the
BV adjusts its speed to regulate the timing of the
interaction.

To examine whether Video2Track can vary
the passing order at a fixed target APET,
we compared the yielding trial from RQ1
with a rushing trial using the same target
APET, road configuration, and reference routes.
As shown in Fig.~\ref{FIG:Cross_PET_closed_track},
the VUT passed through the conflict region
first in the yielding trial, whereas the BV
passed first in the rushing trial shown in
Fig.~\ref{fig:cross_rush}.
The observed minimum APET values were
2.60 and 2.52~s, respectively, differing
by only 0.08~s.
These results show that Video2Track can
produce opposite passing orders on the
same route pair while maintaining a
similar interaction risk level.

\begin{table}[t]
\centering
\caption{Comparison of adversarial testing
capability using mean minimum APET.}
\label{tab:criticalty}
\begin{tabular*}{\columnwidth}
{@{\extracolsep{\fill}}lcc}
\toprule
Method &
\multicolumn{2}{c}{Mean Minimum APET (s)} \\
\cmidrule(lr){2-3}
& Cut-in & Unprotected Left-turn \\
\midrule
Predefined Scenario & 2.64 & 2.54 \\
TroubleMaker & 0.90 & 1.71 \\
\textbf{Video2Track (Ours)}
& \textbf{0.48} & \textbf{0.84} \\
\bottomrule
\end{tabular*}
\end{table}

\begin{figure}[!t]
    \centering

    \subfloat[Predefined Scenario]{%
        \begin{minipage}[t]{\columnwidth}
            \centering
            \includegraphics[
                width=0.48\linewidth,
                height=2.5cm
            ]{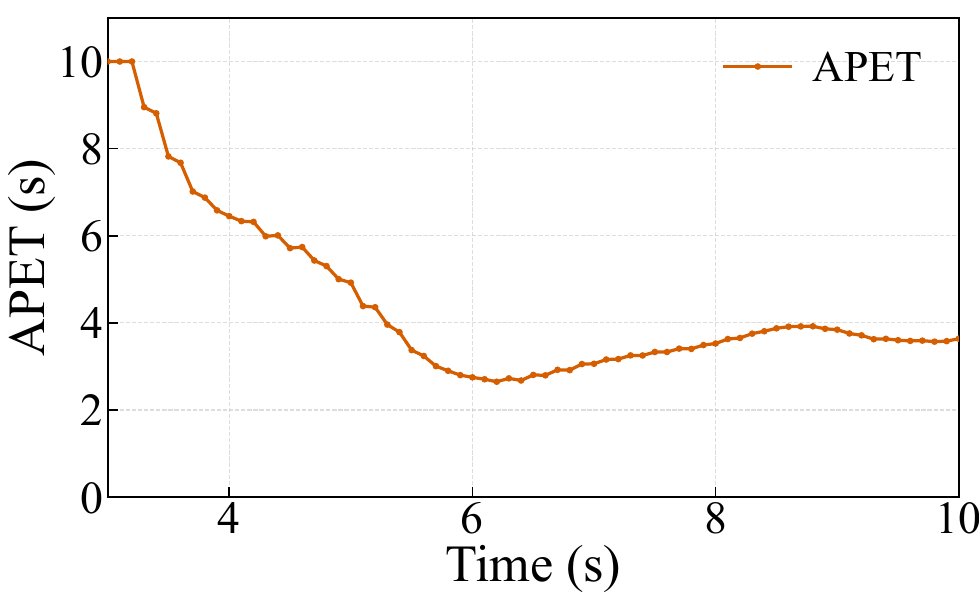}%
            \hfill
            \includegraphics[
                width=0.48\linewidth,
                height=2.5cm
            ]{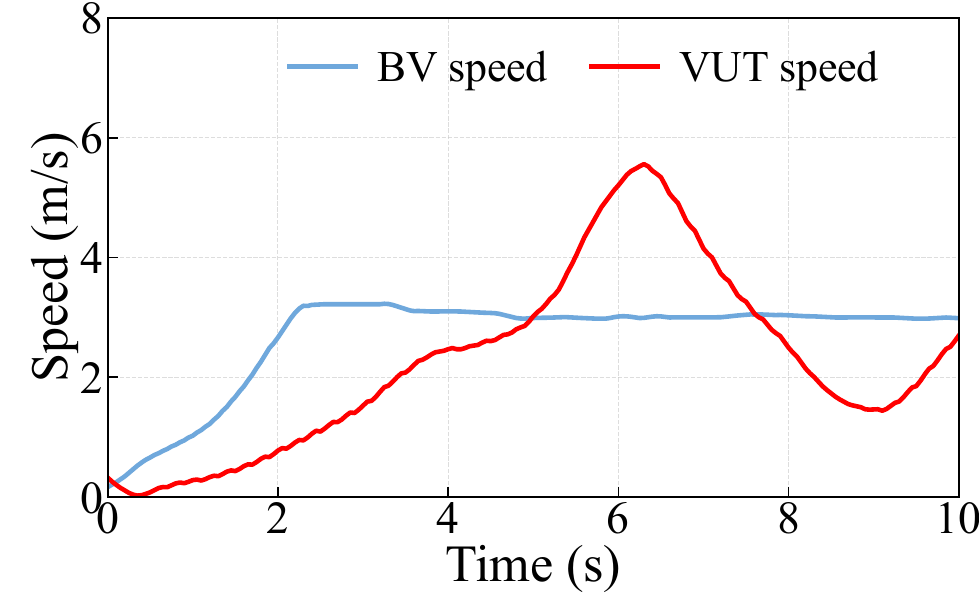}
        \end{minipage}%
        \label{fig:rq3_b1}%
    }

    \par\vspace{0.3em}

    \subfloat[TroubleMaker]{%
        \begin{minipage}[t]{\columnwidth}
            \centering
            \includegraphics[
                width=0.48\linewidth,
                height=2.5cm
            ]{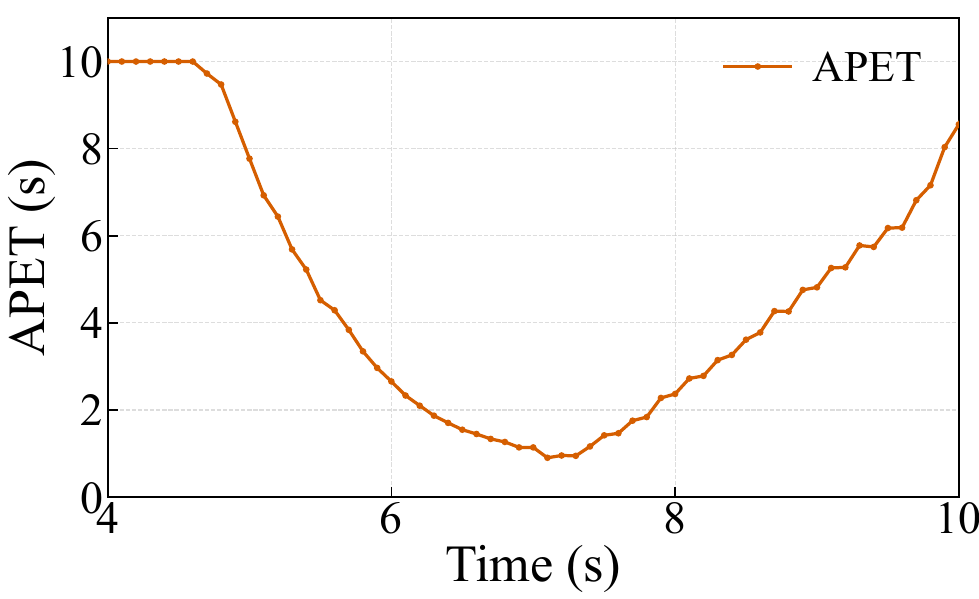}%
            \hfill
            \includegraphics[
                width=0.48\linewidth,
                height=2.5cm
            ]{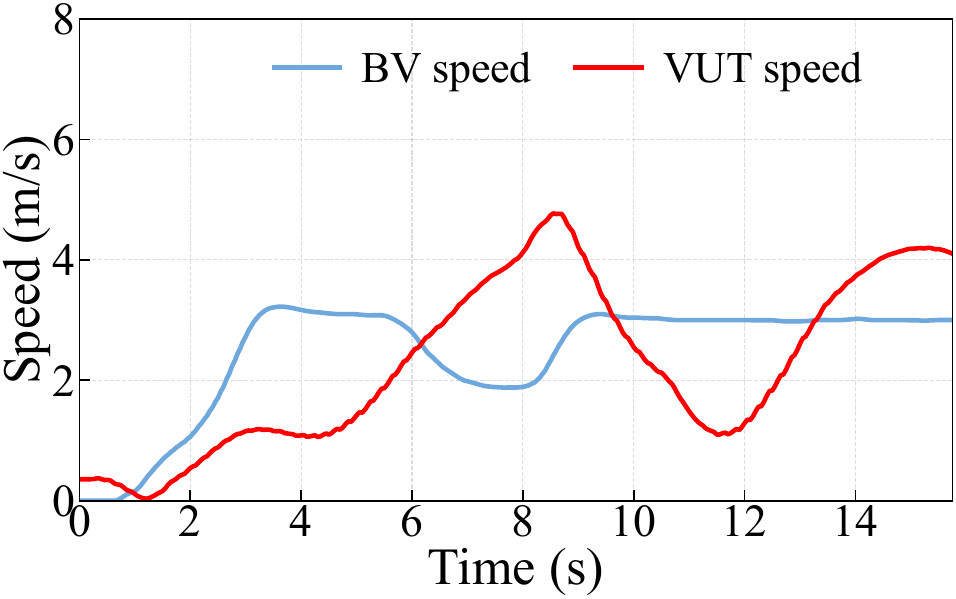}
        \end{minipage}%
        \label{fig:rq3_b2}%
    }

    \par\vspace{0.3em}

    \subfloat[Video2Track]{%
        \begin{minipage}[t]{\columnwidth}
            \centering
            \includegraphics[
                width=0.48\linewidth,
                height=2.5cm
            ]{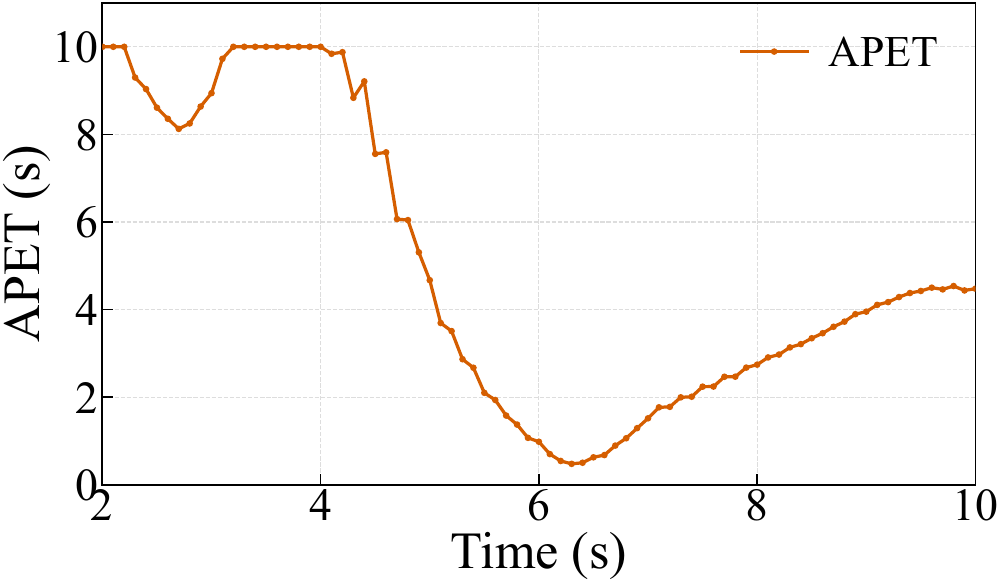}%
            \hfill
            \includegraphics[
                width=0.48\linewidth,
                height=2.5cm
            ]{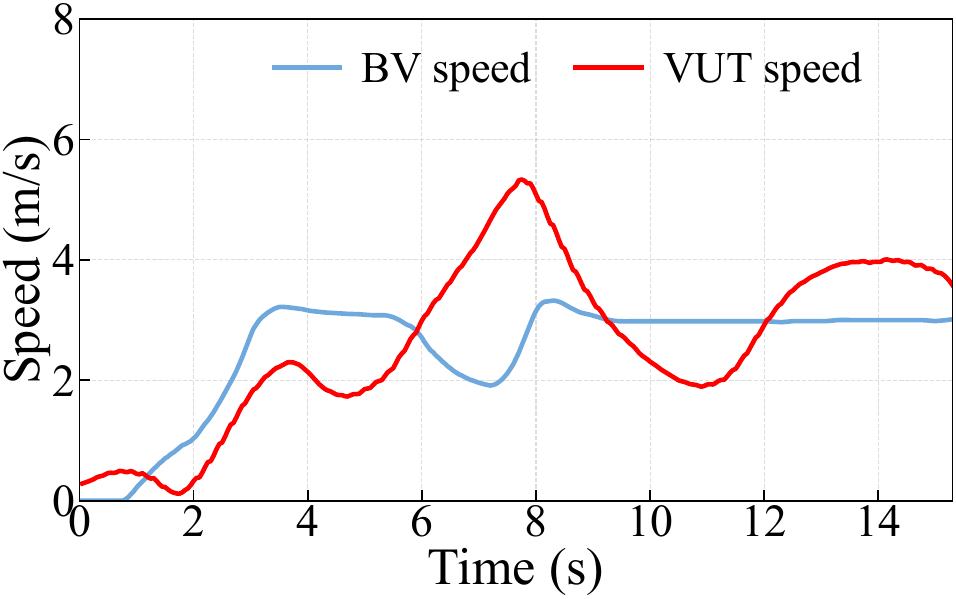}
        \end{minipage}%
        \label{fig:rq3_ours}%
    }

    \vspace{-0.4em}

    \caption{Comparison of representative cut-in
    executions during proving-ground testing.
    Each row shows the APET evolution (left)
    and vehicle speed profiles (right)
    for the corresponding method.}

    \label{fig:rq3_cases}
\end{figure}

\subsubsection{RQ3: Adversarial Testing Capability}
\label{sec:4.2.3}

We further evaluate the ability of Video2Track
to generate temporally critical interactions,
using smaller minimum APET values as an
indicator of greater interaction criticality.
Fig.~\ref{fig:rq3_cases} presents
representative cut-in executions of
the compared methods, including APET
evolution and vehicle speed profiles.
In the Predefined Scenario, the BV
follows a preset speed profile without
adapting to VUT responses, resulting
in relatively large APET values.
By contrast, TroubleMaker and Video2Track
adjust BV behavior online based on
VUT feedback, producing more
temporally critical interactions.

Table~\ref{tab:criticalty} compares
the mean minimum APET of the three
methods across repeated proving-ground
executions.
Video2Track achieves the lowest values
of 0.48~s and 0.84~s in the cut-in
and unprotected left-turn scenarios,
representing reductions of 46.7\%
and 50.9\%, respectively, compared
with TroubleMaker.
These results demonstrate the capability
of Video2Track to generate more
temporally critical interactions
for adversarial proving-ground testing.

\subsection{Parameter Sensitivity Analysis}
\label{sec:4.3}

We investigate the influence of the
adversarial weight on APET regulation
and acceleration demand.
With $\omega_{\mathrm{real}}=1$,
$\omega_{\mathrm{adv}}$ is varied under
two target APET settings, $\tau=2.0$~s
and $\tau=3.0$~s.
The configuration $\omega_{\mathrm{adv}}=0$
serves as the non-adversarial baseline.
Acceleration demand is evaluated using
the 95th-percentile acceleration magnitude.

\begin{figure}[!t]
    \centering
    \vspace{-0.3cm} 

    \subfloat[APET error]{%
        \includegraphics[
            width=0.48\columnwidth,
            height=2.5cm
        ]{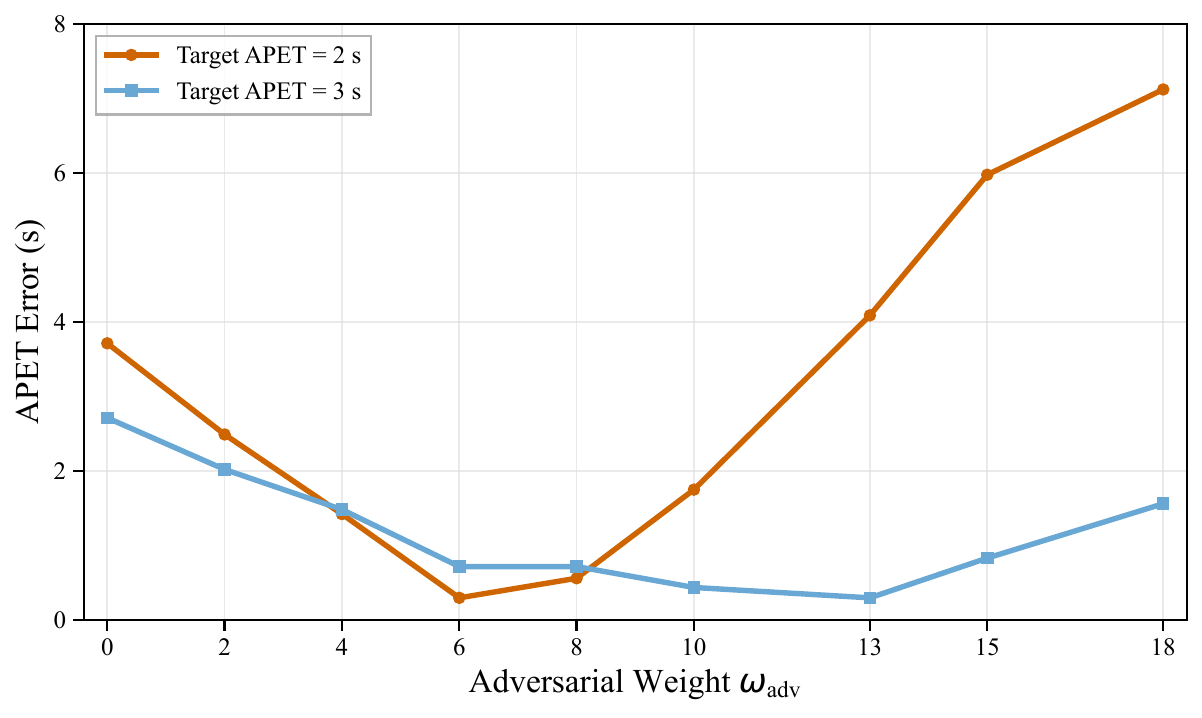}%
        \label{fig:sensitivity_apet}%
    }%
    \hfill
    \subfloat[Acceleration]{%
        \includegraphics[
            width=0.48\columnwidth,
            height=2.5cm
        ]{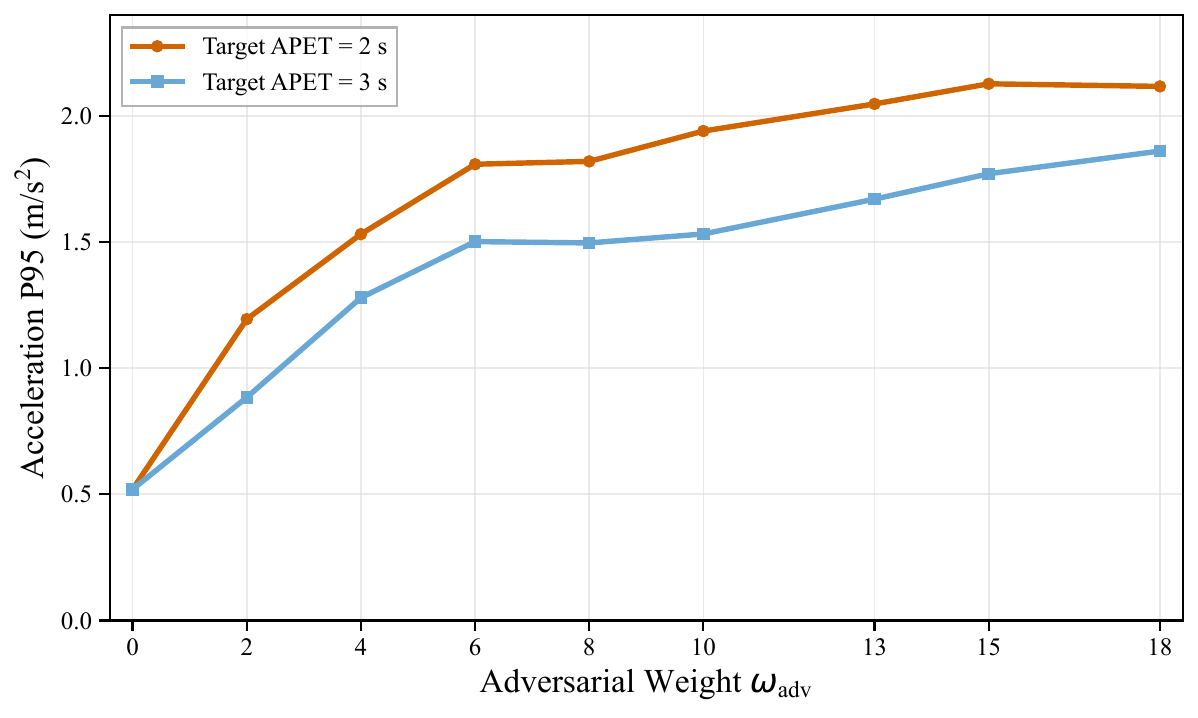}%
        \label{fig:sensitivity_acc}%
    }

    \vspace{-0.4em}

    \caption{Sensitivity of APET regulation
    accuracy and acceleration magnitude
    to the adversarial weight under
    different target APET settings.}

    \label{fig:game_sensitivity}
\end{figure}

As shown in Fig.~\ref{fig:game_sensitivity},
increasing $\omega_{\mathrm{adv}}$
initially reduces the APET error,
while increasing acceleration demand.
The APET error falls below 0.5~s
at $\omega_{\mathrm{adv}}=6$ for
$\tau=2.0$~s and at
$\omega_{\mathrm{adv}}=10$ and $13$
for $\tau=3.0$~s.
At higher weights, the APET error
increases despite the continued rise
in acceleration.
These results show that stronger
adversarial weighting does not
consistently improve APET regulation
and may require more aggressive
vehicle maneuvers.

\section{Conclusion and Future Work}
\label{sec:5}

This paper presented Video2Track, a framework
for transferring real-world driving interactions
from videos to steerable adversarial scenarios
for proving-ground testing of automated
driving systems.
By combining interaction-aware maneuver-pair
retrieval with guided diffusion, Video2Track
achieved an ICR of 82.1\% and an SGSR of 78.6\%.
During closed-track testing, game-based online
refinement adjusted BV trajectories based on
real-time VUT feedback, with the mean minimum
APET remaining within 0.5~s of all four targets.
Compared with the evaluated baselines,
Video2Track achieved lower mean minimum APET
values in both cut-in and unprotected left-turn
scenarios, producing more temporally
critical interactions.
The results demonstrate the use of
video-derived interactions for steerable
adversarial testing on a 5G cloud-controlled
proving ground.

Future work will extend Video2Track to a wider
range of interactions, including scenarios involving
multiple background vehicles. We will also
investigate closer integration of diffusion-based
scenario generation and game-based online refinement
to improve computational efficiency and adaptability
to different responses from the vehicle under test.








\bibliographystyle{IEEEtran}
\bibliography{references}

@ARTICLE{zhang2025real,
  author={Zhang, Xinrui and Xiong, Lu and Zhang, Peizhi and Huang, Junpeng and Ma, Yining},
  title={Real-World Troublemaker: A 5G Cloud-Controlled Track Testing Framework for Automated Driving Systems in Safety-Critical Interaction Scenarios},
  journal={IEEE Internet of Things Journal},
  volume  = {12},
  year={2025},
  pages= {50617-50631}
}

@inproceedings{tian2025cloud,
  title={Cloud-based Predictive Path Tracking Control for Global Vehicle Targets with Uncertain Latency},
  author={Tian, Mengjie and Zhang, Peizhi and Zhuo, Guirong and Zhang, Xinrui and Ma, Yining and Wang, Xiurong and Xiong, Lu},
  booktitle={2025 IEEE Intelligent Vehicles Symposium (IV)},
  year={2025},
  pages={1654--1660},

}

@article{liu2024curse,
  title={Curse of rarity for autonomous vehicles},
  author={Liu, Henry X and Feng, Shuo},
  journal={nature communications},
  volume={15},
  year={2024},
  pages={4808}
}

@article{wang2022autonomous,
  title={Autonomous driving testing scenario generation based on in-depth vehicle-to-powered two-wheeler crash data in China},
  author={Wang, Xinghua and Peng, Yong and Xu, Tuo and Xu, Qian and Wu, Xianhui and Xiang, Guoliang and Yi, Shengen and Wang, Honggang},
  journal={Accident Analysis \& Prevention},
  volume={176},
  year={2022},
  pages={106812}
}

@article{feng2020safety,
  title={Safety assessment of highly automated driving systems in test tracks: A new framework},
  author={Feng, Shuo and Feng, Yiheng and Yan, Xintao and Shen, Shengyin and Xu, Shaobing and Liu, Henry X},
  journal={Accident Analysis \& Prevention},
  volume={144},
  year={2020},
  pages={105664}
}

@article{zhang2025testing,
  title   = {Testing through conflict: Cloud-coordinated closed-field testing for multi-agent continuous interaction scenarios},
  author  = {Xinrui Zhang and Lu Xiong and Peizhi Zhang and Haojie Feng and Junpeng Huang and Mengjie Tian},
  journal = {Chinese Journal of Mechanical Engineering},
  volume  = {39},
  pages   = {100192},
  year    = {2026},
  doi     = {10.1016/j.cjme.2025.100192}
}

@article{wu2025evolving,
  title={An Evolving Scenario Generation Method based on Dual-modal Driver Model Trained by Multi-Agent Reinforcement Learning},
  author={Wu, Xinzheng and Chen, Junyi and Ye, Shaolingfeng and Jiang, Wei and Shen, Yong},
  journal={arXiv preprint arXiv:2508.02027},
  year={2025}
}

@inproceedings{liu2025isfm4sim,
  title={ISFM4Sim: A Two-Wheeler Driving Behavior Model for Simulation Testing of Autonomous Vehicles},
  author={Liu, Zhenyuan and Wang, Qiyi and Zhang, Longgao and Qin, Jiao and Wang, Junjie and Chen, Junyi and Wang, Xuesong},
  booktitle={2025 IEEE Intelligent Vehicles Symposium (IV)},
  year={2025},
  pages={2525--2531}
}

@inproceedings{xu2025diffscene,
  author={Xu, Chejian and Petiushko, Aleksandr and Zhao, Ding and Li, Bo},
  title={Diffscene: Diffusion-based safety-critical scenario generation for autonomous vehicles},
  booktitle={Proceedings of the AAAI conference on artificial intelligence},
  volume={39},
  year={2025},
  pages={8797--8805},
}

@article{lu2025omnitester,
  title={OmniTester: Multimodal Large Language Model Driven Scenario Testing for Autonomous Vehicles},
  author={Lu, Qiujing and Wang, Xuanhan and Jiang, Yiwei and Zhao, Guangming and Ma, Mingyue and Feng, Shuo},
  journal={Automotive Innovation},
  volume={8},
  pages={838--852},
  year={2025},
}

@inproceedings{zhang2025drivegen,
  title={Drivegen: Towards infinite diverse traffic scenarios with large models},
  author={Zhang, Shenyu and Tian, Jiaguo and Zhu, Zhengbang and Huang, Shan and Yang, Jucheng and Zhang, Weinan},
  booktitle={2025 IEEE/RSJ International Conference on Intelligent Robots and Systems (IROS)},
  pages={10100--10107},
  year={2025},
}

@techreport{iso2023road,
  author      = {{International Organization for Standardization}},
  title       = {{Road vehicles --- Test scenarios for
                  automated driving systems ---
                  Scenario based safety evaluation framework}},
  institution = {International Organization for Standardization},
  type        = {Standard},
  number      = {ISO 34502:2022(E)},
  address     = {Geneva, Switzerland},
  year        = {2022}
}

@inproceedings{fremont2019scenic,
  title={Scenic: a language for scenario specification and scene generation},
  author={Fremont, Daniel J and Dreossi, Tommaso and Ghosh, Shromona and Yue, Xiangyu and Sangiovanni-Vincentelli, Alberto L and Seshia, Sanjit A},
  booktitle={Proceedings of the 40th ACM SIGPLAN conference on programming language design and implementation},
  pages={63--78},
  year={2019}
}

@incollection{winner2018pegasus,
  title={PEGASUS—First steps for the safe introduction of automated driving},
  author={Winner, Hermann and Lemmer, Karsten and Form, Thomas and Mazzega, Jens},
  booktitle={Road Vehicle Automation 5},
  pages={185--195},
  year={2018},
}

@article{xia2018test,
  title={Test scenario design for intelligent driving system ensuring coverage and effectiveness},
  author={Xia, Qin and Duan, Jianli and Gao, Feng and Hu, Qiuxia and He, Yingdong},
  journal={International Journal of Automotive Technology},
  volume={19},
  pages={751--758},
  year={2018},
}

@article{miao2024dashcam,
  title={From dashcam videos to driving simulations: Stress testing automated vehicles against rare events},
  author={Miao, Yan and Fainekos, Georgios and Hoxha, Bardh and Okamoto, Hideki and Prokhorov, Danil and Mitra, Sayan},
  journal={arXiv preprint arXiv:2411.16027},
  year={2024}
}

@INPROCEEDINGS{trafficgen,
  author={Feng, Lan and Li, Quanyi and Peng, Zhenghao and Tan, Shuhan and Zhou, Bolei},
  booktitle={2023 IEEE International Conference on Robotics and Automation (ICRA)}, 
  title={TrafficGen: Learning to Generate Diverse and Realistic Traffic Scenarios}, 
  year={2023},
  volume={},
  number={},
  pages={3567-3575},
}

@ARTICLE{dingzhaosurvy,
  author={Ding, Wenhao and Xu, Chejian and Arief, Mansur and Lin, Haohong and Li, Bo and Zhao, Ding},
  journal={IEEE Transactions on Intelligent Transportation Systems}, 
  title={A Survey on Safety-Critical Driving Scenario Generation—A Methodological Perspective}, 
  year={2023},
  volume={24},
  pages={6971-6988}
}

@INPROCEEDINGS{ctg2023,
  author={Zhong, Ziyuan and Rempe, Davis and Xu, Danfei and Chen, Yuxiao and Veer, Sushant and Che, Tong and Ray, Baishakhi and Pavone, Marco},
  booktitle={2023 IEEE International Conference on Robotics and Automation (ICRA)}, 
  title={Guided Conditional Diffusion for Controllable Traffic Simulation}, 
  year={2023},
  volume={},
  number={},
  pages={3560-3566},
}

@article{kullgren2010comparison,
  title={Comparison between Euro NCAP test results and real-world crash data},
  author={Kullgren, Anders and Lie, Anders and Tingvall, Claes},
  journal={Traffic injury prevention},
  volume={11},
  pages={587--593},
  year={2010},
}

@article{sun2021scenario,
  title={Scenario-based test automation for highly automated vehicles: A review and paving the way for systematic safety assurance},
  author={Sun, Jian and Zhang, He and Zhou, Huajun and Yu, Rongjie and Tian, Ye},
  journal={IEEE transactions on intelligent transportation systems},
  volume={23},
  pages={14088--14103},
  year={2021},
}

@article{feng2023dense,
  title={Dense reinforcement learning for safety validation of autonomous vehicles},
  author={Feng, Shuo and Sun, Haowei and Yan, Xintao and Zhu, Haojie and Zou, Zhengxia and Shen, Shengyin and Liu, Henry X},
  journal={Nature},
  volume={615},
  pages={620--627},
  year={2023}
}

@ARTICLE{liTITS2026,
  author={Li, Zhuoren and Leng, Bo and Xiong, Lu and  Eichberger, Arno and Huang, Chao and Hu, Jia},
  journal={IEEE Transactions on Intelligent Transportation Systems}, 
  title={Safety-Enhanced Deep Reinforcement Learning for Autonomous Driving: Dare to Make Mistakes to Learn Better and Faster}, 
  year={2026},
  volume={},
  number={}
}

@inproceedings{
liuvlm,
title={{VLM}-Enhanced Adversarial Scene Generation from Images and Videos for Safe Autonomous Driving},
author={Tianyi Liu and Yingjie Xu and Yinlong Liu},
booktitle={Embodied and Safe-Assured Robotic Systems},
year={2025}
}

@ARTICLE{11457031,
  author={Jin, Guizhe and Li, Zhuoren and Leng, Bo and Han, Wei and Xiong, Lu and Sun, Chen},
  journal={IEEE Transactions on Neural Networks and Learning Systems}, 
  title={Hybrid Action-Based Reinforcement Learning for Multiobjective Compatible Autonomous Driving}, 
  year={2026}}

@ARTICLE{Zhangzq,
  author={Zhang, Qixiang and Xing, Yang and Wang, Jinxiang and Fang, Zhenwu and Liu, Yahui and Yin, Guodong},
  journal={IEEE Transactions on Intelligent Transportation Systems}, 
  title={Interaction-Aware and Driving Style-Aware Trajectory Prediction for Heterogeneous Vehicles in Mixed Traffic Environment}, 
  year={2025},
  volume={26},
  pages={10710-10724}}

@misc{videoscenario_hf,
  author       = {{Jay1101}},
  title        = {{VideoScenario}},
  year         = {2026},
  howpublished = {Hugging Face Dataset},
  url          = {https://huggingface.co/datasets/Jay1101/VideoScenario},
  note         = {Accessed: Nov. 23, 2026}
}

@ARTICLE{vbd,
  author={Huang, Zhiyu and Zhang, Zixu and Vaidya, Ameya and Chen, Yuxiao and Fernández Fisac, Jaime and Lv, Chen},
  journal={IEEE Transactions on Intelligent Transportation Systems}, 
  title={Versatile Behavior Diffusion for Generalized Traffic Agent Simulation}, 
  year={2026},
  volume={27},
  number={7},
  pages={8681-8697},
  doi={10.1109/TITS.2026.3662886}}

@article{WANG202417,
  author = {Wang, Hong and Shao, Wenbo
            and Sun, Chen and Yang, Kai
            and Cao, Dongpu and Li, Jun},
  title = {A Survey on an Emerging Safety Challenge
           for Autonomous Vehicles:
           Safety of the Intended Functionality},
  journal = {Engineering},
  volume = {33},
  pages = {17--34},
  year = {2024},
  doi = {10.1016/j.eng.2023.10.011}
}

@ARTICLE{zhu2026Iotj,
  author={Zhu, Bing and Huang, Yinzi and Zhao, Jian and Zhang, Peixing and Li, Wenxu and Gao, Ming and Fan, Xinwei},
  journal={IEEE Internet of Things Journal}, 
  title={A Systematic Review of Influences and Test Methodologies: Evaluating the Impact of Adverse Weather Conditions on Perception Systems for Intelligent Vehicles}, 
  year={2026},
  volume={13},
  number={19},
  pages={44100-44123},
  doi={10.1109/JIOT.2026.3705957}}

@article{zhangpeixing2023,
author = {Peixing Zhang and Bing Zhu and Jian Zhao and Tianxin Fan and Yuhang Sun},
title = {Safety evaluation method in multi-logical scenarios for automated vehicles based on naturalistic driving trajectory},
journal = {Accident Analysis \& Prevention},
volume = {180},
pages = {106926},
year = {2023},
issn = {0001-4575},
doi = {https://doi.org/10.1016/j.aap.2022.106926},
}

@article{li2026mrisce,
title = {MRISce: an interactive autonomous driving test scenario generation method based on multi-agent reinforcement learning},
journal = {Accident Analysis \& Prevention},
volume = {234},
pages = {108611},
year = {2026},
issn = {0001-4575},
doi = {https://doi.org/10.1016/j.aap.2026.108611},
author = {Jiwei Li and Runmin Wang and Yu Zhu and Mengying Zhang and Xiangmo Zhao},
}

@article{yang2021cavtest,
  author = {Lan Yang and Runmin Wang
            and Xiangmo Zhao and Zhigang Xu
            and Yipeng Yang},
  title = {CAVTest: A Closed Connected
           and Automated Vehicles Test Field
           of Chang'an University in China},
  journal = {SAE International Journal of
             Connected and Automated Vehicles},
  volume = {4},
  number = {4},
  pages = {423--435},
  year = {2021},
  doi = {10.4271/12-04-04-0032}
}

@article{ma2024evolving,
title = {Evolving testing scenario generation and intelligence evaluation for automated vehicles},
journal = {Transportation Research Part C: Emerging Technologies},
volume = {163},
pages = {104620},
year = {2024},
issn = {0968-090X},
doi = {https://doi.org/10.1016/j.trc.2024.104620},
author = {Yining Ma and Wei Jiang and Lingtong Zhang and Junyi Chen and Hong Wang and Chen Lv and Xuesong Wang and Lu Xiong},
}

@inproceedings{
cannons2026from,
title={From Segments to Scenes: Temporal Understanding for Agentic Autonomous Driving via Vision-Language Models},
author={Kevin Cannons and Saeed Ranjbar Alvar and Mohammad Asiful Hossain and Ahmad Rezaei and Mohsen Gholami and Alireza Heidari and Zhou Weimin and Yong Zhang and Mohammad Akbari},
booktitle={Second Workshop on Agents in the Wild: Safety, Security, and Beyond},
year={2026},
url={https://openreview.net/forum?id=ZTnt0PzHnL}
}

@ARTICLE{chen2021how,
  author={Chen, Rui and Arief, Mansur and Zhang, Weiyang and Zhao, Ding},
  journal={IEEE Transactions on Intelligent Transportation Systems}, 
  title={How to Evaluate Proving Grounds for Self-Driving? A Quantitative Approach}, 
  year={2021},
  volume={22},
  number={9},
  pages={5737-5748},
  doi={10.1109/TITS.2020.2991757}
  }

@misc{chat2scenic,
      title={Chat2Scenic: An Iterative RAG-Based Framework for Scenario Generation in Autonomous Driving}, 
      author={Yuan Gao and Wenting Miao and Mattia Piccinini and Haoyu Wang and Qunying Song and Johannes Betz},
      year={2026},
      eprint={2607.14387},
      archivePrefix={arXiv},
      primaryClass={cs.AI},
      url={https://arxiv.org/abs/2607.14387}, 
}

@INPROCEEDINGS{chatscene,
  author={Zhang, Jiawei and Xu, Chejian and Li, Bo},
  booktitle={2024 IEEE/CVF Conference on Computer Vision and Pattern Recognition (CVPR)}, 
  title={ChatScene: Knowledge-Enabled Safety-Critical Scenario Generation for Autonomous Vehicles}, 
  year={2024},
  volume={},
  number={},
  pages={15459-15469},
  doi={10.1109/CVPR52733.2024.01464}}

@ARTICLE{queiroz2024driver,
  author={Queiroz, Rodrigo and Sharma, Divit and Caldas, Ricardo and Czarnecki, Krzysztof and García, Sergio and Berger, Thorsten and Pelliccione, Patrizio},
  journal={IEEE Transactions on Intelligent Transportation Systems}, 
  title={A Driver-Vehicle Model for ADS Scenario-Based Testing}, 
  year={2024},
  volume={25},
  number={8},
  pages={8641-8654},
  doi={10.1109/TITS.2024.3373531}}

@techreport{euroncap2024aeb,
  author      = {{Euro NCAP}},
  title       = {{Test Protocol -- AEB Car-to-Car Systems}},
  institution = {European New Car Assessment Programme},
  year        = {2024},
  number      = {Version 4.3.1},
  url         = {https://cdn.euroncap.com/cars/assets/euro_ncap_aeb_c2c_test_protocol_v431_532926aad1.pdf}
}

\end{document}